\documentclass[pdflatex, sn-standardnature]{sn-jnl}

\jyear{2022}%

\theoremstyle{thmstyleone}%
\theoremstyle{thmstyletwo}%
\theoremstyle{thmstylethree}%
\begin{document}

\title[Exploring Robust Architectures for Deep Artificial Neural Networks]{Exploring Robust Architectures for Deep Artificial Neural Networks}

\author*[1]{\fnm{Asim} \sur{Waqas}}\email{waqasa8@students.rowan.edu}

\author[2]{\fnm{Hamza} \sur{Farooq}}\email{faroo014@umn.edu}

\author[1]{\fnm{Nidhal C.} \sur{Bouaynaya}}\email{bouaynaya@rowan.edu}

\author[1]{\fnm{Ghulam} \sur{Rasool}}\email{rasool@rowan.edu}

\affil*[1]{\orgdiv{Department of Electrical and Computer Engineering}, \orgname{Rowan University}, \orgaddress{\city{Glassboro}, \postcode{08028}, \state{NJ}, \country{USA}}}

\affil[2]{\orgdiv{Department of Radiology}, \orgname{University of Minnesota}, \orgaddress{\city{Minneapolis}, \postcode{55455}, \state{MN}, \country{USA}}}

\abstract{
The architectures of deep artificial neural networks (DANNs) are routinely studied to improve their predictive performance. However, the relationship between the architecture of a DANN and its robustness to noise and adversarial attacks is less explored. We investigate how the robustness of DANNs relates to their underlying graph architectures or structures. This study: (1) starts by exploring the design space of architectures of DANNs using graph-theoretic robustness measures; (2) transforms the graphs to DANN architectures to train/validate/test on various image classification tasks; (3) explores the relationship between the robustness of trained DANNs against noise and adversarial attacks and the robustness of their underlying architectures estimated via graph-theoretic measures. We show that the topological entropy and Olivier-Ricci curvature of the underlying graphs can quantify the robustness performance of DANNs. The said relationship is stronger for complex tasks and large DANNs. Our work will allow autoML and neural architecture search community to explore design spaces of robust and accurate DANNs.
}

\keywords{Deep artificial neural networks, Graph theory, Curvature, Entropy, Adversarial robustness}


\maketitle

\section{Introduction}\label{sec1}
The architecture or structure of a deep artificial neural network (DANN) is defined by the connectivity patterns among its constituent artificial neurons. The mere presence or absence of a connection between two neurons or a set of neurons may provide a useful prior and improve the predictive performance of a DANN. A range of architectures has been developed over years to tackle various machine learning tasks in computer vision, natural language processing, and reinforcement learning \cite{DBLP:conf/cvpr/HeZRS16, DBLP:conf/icml/SaxeKCBSN11, lecun1998gradient, krizhevsky2017imagenet, DBLP:conf/cvpr/SzegedyLJSRAEVR15}. In general, the process of the development of DANN architectures is manual, iterative and time consuming. AutoML and neural architecture search (NAS) attempt to use machine learning and search the design space of DANNs for architectures that may yield maximum test data accuracy. After the selection of a suitable DANN architecture for the given task, the optimal values of the connections (parameters or weights) are found using the training dataset and the well-known gradient descent or one of its variant algorithms. Recently, considerable research efforts have been focused on automating the laborious task of DANN architecture design and development using techniques of autoML and NAS. However, all such efforts are primarily focused on improving the test accuracy of the DANN on the given task.

In the real world, DANNs face the challenging problem of maintaining their predictive performance in the face of uncertainties and noise in the input data \cite{dera2021premium}. The challenge is further exacerbated for mission-critical application areas, such as clinical diagnosis, autonomous driving, financial decision-making, and defense. Ideally, a real world deployment-ready DANN should be robust to or equivalently maintain its predictive performance against two different types of noise, natural and malicious. The natural noise is related to the out-of-distribution generalization. Such noise is caused by the day-to-day changes in input data, e.g., changes in the hardware or software configurations used for processing input data. The malicious or adversarial noise is imperceptible to human eye and is generated by an adversary for fooling the DANN into producing an erroneous decision \cite{DBLP:journals/corr/SzegedyZSBEGF13}.

It has been shown with the help of Percolation theory that the architecture or structure underlying a network of any real-world system may play a key role in defining its robustness to various insults and attacks \cite{barabasi2016network}. Graph-theoretic measures, such as network topological entropy and Ollivier-Ricci curvature, successfully quantify the functional robustness of various networks \cite{tannenbaum2015ricci}. Examples include studying the behavior of cancer cells, analyzing the fragility of financial networks, studying the robustness of brain networks, tracking changes attributable to age and Autism Spectrum Disorder (ASD), and explaining cognitive impairment in Multiple Sclerosis (MS) patients \cite{sandhu2015graph, sandhu2016ricci, farooq2019network, farooq2020robustness}. Recently, the relationship between the architectures of DANNs (quantified by various graph-theoretic measures before training) and their predictive accuracy (available after training) has been established \cite{DBLP:conf/iccv/XieKGH19, DBLP:conf/icml/YouLHX20}. Various graph-theoretic measures (e.g., path length and clustering coefficient) calculated from the architectures of DANNs are quantitatively linked to their accuracy on various image classification tasks. However, the relationship between the graph-theoretic measures related to the robustness (entropy and Ollivier-Ricci curvature) of the architecture of DANNs and their performance against natural and adversarial noise has never been explored. Establishing such a relationship will allow the autoML and NAS research community to design and develop robust DANNs without training and testing these architectures. 

In this work, we study graph-theoretic properties of architectures of DANNs to quantify their test-time robustness. Specifically, we use the graph measures of topological entropy and curvature of the architecture of DANNs as robustness metrics. We make two distinct research contributions to the robustness analysis of DANNs: (1) We establish a quantitative relationship between the graph-theoretic robustness measures of entropy and curvature of DANNs (available before training) and the robustness of these DANNs to natural and adversarial noise (evaluated after training DANNs). Previous studies explored graph measures that relate to the performance of DANNs, but robustness of DANNs through graph-robustness measures has never been studied. We show that graph entropy and curvature are related to DANNs' robustness and these structural measures can identify robust architectures of DANNs even before training for the given task. (2) We show that relationship between the graph robustness measured using entropy and Ollivier-Ricci curvature and the robustness performance of DANN against noise and adversarial attacks becomes significantly stronger for complex tasks, larger datasets, and bigger DANNs. Given that the sizes of DANNs and the complexity of tasks/datasets are growing significantly for many real-world applications, the strong entropy-robustness relationship assumes greater importance. The autoML/NAS design problems where robustness of DANNs is vital, our analysis can help identify robust architectures without the need to train and test these DANNs under various noisy conditions.

  \begin{figure}[ht]%
    \centering
    \includegraphics[width=1\textwidth]{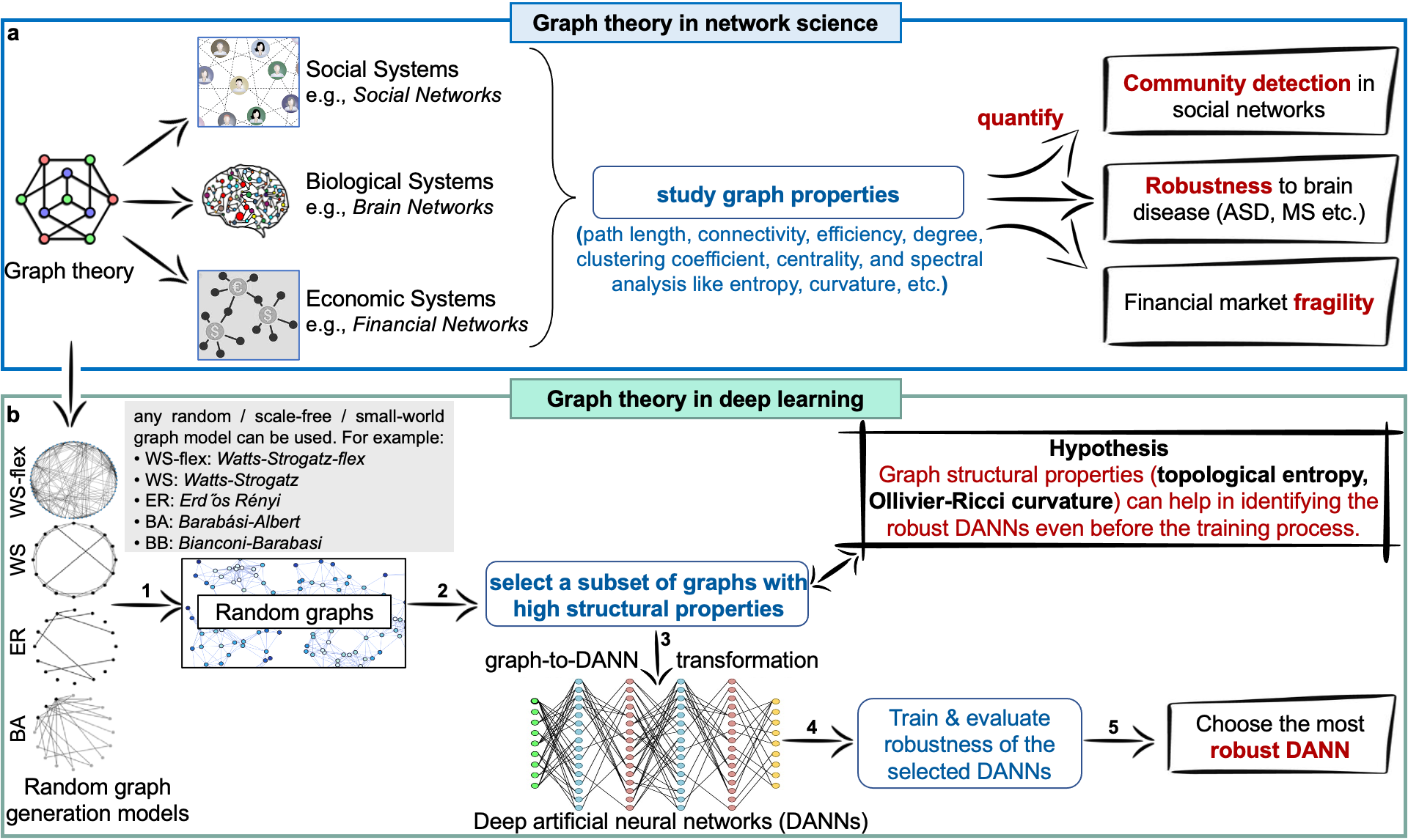}
    \caption{Exploring robustness of deep artificial neural networks (DANNs) with graph-theoretic measures. \textbf{(a)} In the network science (NetSci), real-world systems such as brain networks, financial networks, and social networks are studied using graph-theoretic measures to quantify their robustness and fragility. \textbf{(b)} We use graph-theoretic measures established in NetSci to study graphs of architectures of DANNs. Our approach consists of five steps: (1) build random graphs using classical families of graphs, including ER, BA, WS, and WS-flex, (2) calculate graph-theoretic measures of these random graphs and select a small subset from the entire design space for further analysis, (3) convert selected random graphs into architectures of DANNs, (4) train, validate and test these DANNs under different natural noise and adversarial conditions, and (5) analyze and link the robustness of architectures (measured with graph-theoretic properties) to the performance of trained DANNs against natural noise and adversarial attacks. In summary, using graph-theoretic robustness measures, we can find robust architectures for DANNs without exhaustively training and evaluating many DANNs.}\label{fig:hypothesis}
    \end{figure}

In Fig. \ref{fig:hypothesis}, we provide an overview of the proposed approach. Fig. \ref{fig:hypothesis}(a) illustrates how graph-theoretic measures are often applied in Network Science (NetSci) to study various real-world networks. The illustrated examples include biological systems such as brain networks, economic systems such as financial networks, and social systems such as social networks. Path length, graph connectivity, efficiency, degree measures, clustering coefficient, centrality, and spectral measures (curvature, entropy) are the graph-theoretic measures that researchers have employed for studying real-world networks \cite{sandhu2015graph, farooq2019network, farooq2020robustness, sandhu2016ricci, sia2019ollivier}.\\ Fig.~\ref{fig:hypothesis}(b) illustrates our proposed methodology. We start with building random, scale-free, or small-world networks (or graphs) that are later transformed into architectures of DANNs. We study various graph-theoretic properties of these networks in the graph domain and later quantitatively relate these measures to the robustness of the trained DANNs built from these graphs. We hypothesize that the graph-theoretic measures that quantify the robustness of networks/graphs in the NetSci domain will also provide insight into the robustness of DANNs in the deep learning domain. We provide empirical evidence to support our hypothesis. We use the term \emph{DANN} for deep artificial neural networks, \emph{graphs} for unweighted directed acyclic graphs, and \emph{network} for various networks as used in the network science (NetSci) domain.
\backmatter

\section{Results}\label{sec2}
    \subsection{Graph design space}\label{subsec21}
\begin{figure}[ht]%
        \centering
        \includegraphics[width=0.95\textwidth]{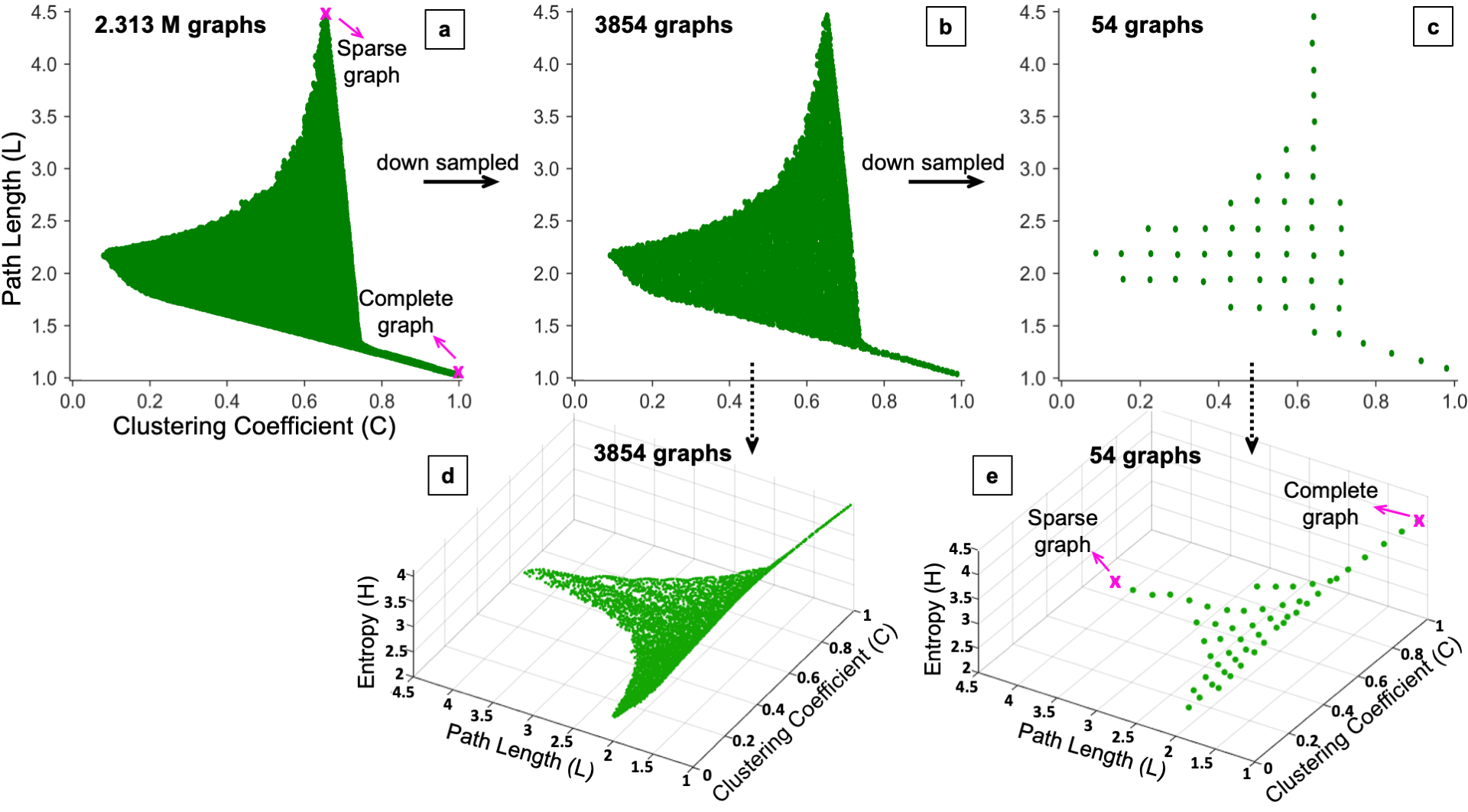}
        \caption{The graph design space for generating random graphs. \textbf{(a),(b),(c):} show 2.313 M candidate graphs from WS-Flex generator downsampled to 3854 and 54 graphs. \textbf{(d)} and \textbf{(e)} show 3854 and 54 graphs in a 3-D space spanned by clustering coefficient (C), path length (L), and entropy (H). Samples of the complete and sparse graphs are identified in (a) and (e).}\label{fig:design-space}
        \end{figure}

    We use two graph measures, \emph{average path length} (L) and \emph{clustering coefficient} (C), for exploring the graph design space. Extensively used in prior works \cite{watts1998collective, sporns2003graph, bassett2006small}, these measures smoothly span the whole design space of the random graphs as illustrated in Fig.~\ref{fig:design-space}. We generate 2.313 Million (M) candidate random graphs using Watts-Strogatz flex (WS-flex) graph generator for a range of C and L values as illustrated in Fig. \ref{fig:design-space}(a). We chose WS-flex because its graphs are superset of graphs generated by three classical methods including, Watts-Strogatz (WS), Erd{\H{o}}s R{\'e}nyi (ER), and Barab{\'a}si-Albert (BA) \cite{watts1998collective, erdHos1960evolution, albert2002statistical}. We downsample 2.313 M candidate WS-flex graphs into coarser bins of 3854 and 54 graphs (Fig.~\ref{fig:design-space}(b)\&\ref{fig:design-space}(c)), where each bin has at least one representative graph. We visualize our candidate graphs using their average path length (L) clustering coefficient (C) and entropy (H), which is a graph-theoretic measure for robustness as show in  Fig.~\ref{fig:design-space}(d)\&\ref{fig:design-space}(e). Fig.~\ref{fig:design-space}(a)\&\ref{fig:design-space}(e) also depict the extreme cases of \emph{complete} and \emph{sparse} graphs. For a complete graph, we have $(C, L, H)=(1.0, 1.0, 4.1)$.
    
    \subsection{From graphs to DANN architectures}\label{subsec22}
    We transform the downsampled 54 graphs into DANNs using the technique of relational graphs proposed by You \emph{et al.} \cite{DBLP:conf/icml/YouLHX20}. We transform the same 54 graphs into multiple types of DANNs including, multilayer perceptrons (MLPs), convolutional neural networks (CNNs), and residual neural networks (ResNets). We use four image classification datasets of varying complexity to train and evaluate DANNs built using 54 different graph structures. These datasets include CIFAR-10, CIFAR-100, Tiny ImageNet, and ImageNet \cite{krizhevsky2009learning, TinyImageNet, DBLP:journals/ijcv/RussakovskyDSKS15}.
    
    The robustness of trained DANNs is quantified by subjecting these models to various levels and types of natural and malicious noise. We used three types of additive noise, Gaussian, Speckle, and Salt\&Pepper. For malicious noise, we employ three different adversarial attacks with varying severity levels. These include Fast Gradient Sign Method (FGSM) \cite{goodfellow2014explaining}, Projected Gradient Descent (PGD) \cite{madry2017towards}, and Carlini Wagner (CW) \cite{DBLP:conf/sp/Carlini017}.
    
    \begin{figure}[ht]%
        \centering
        \includegraphics[width=0.98\textwidth]{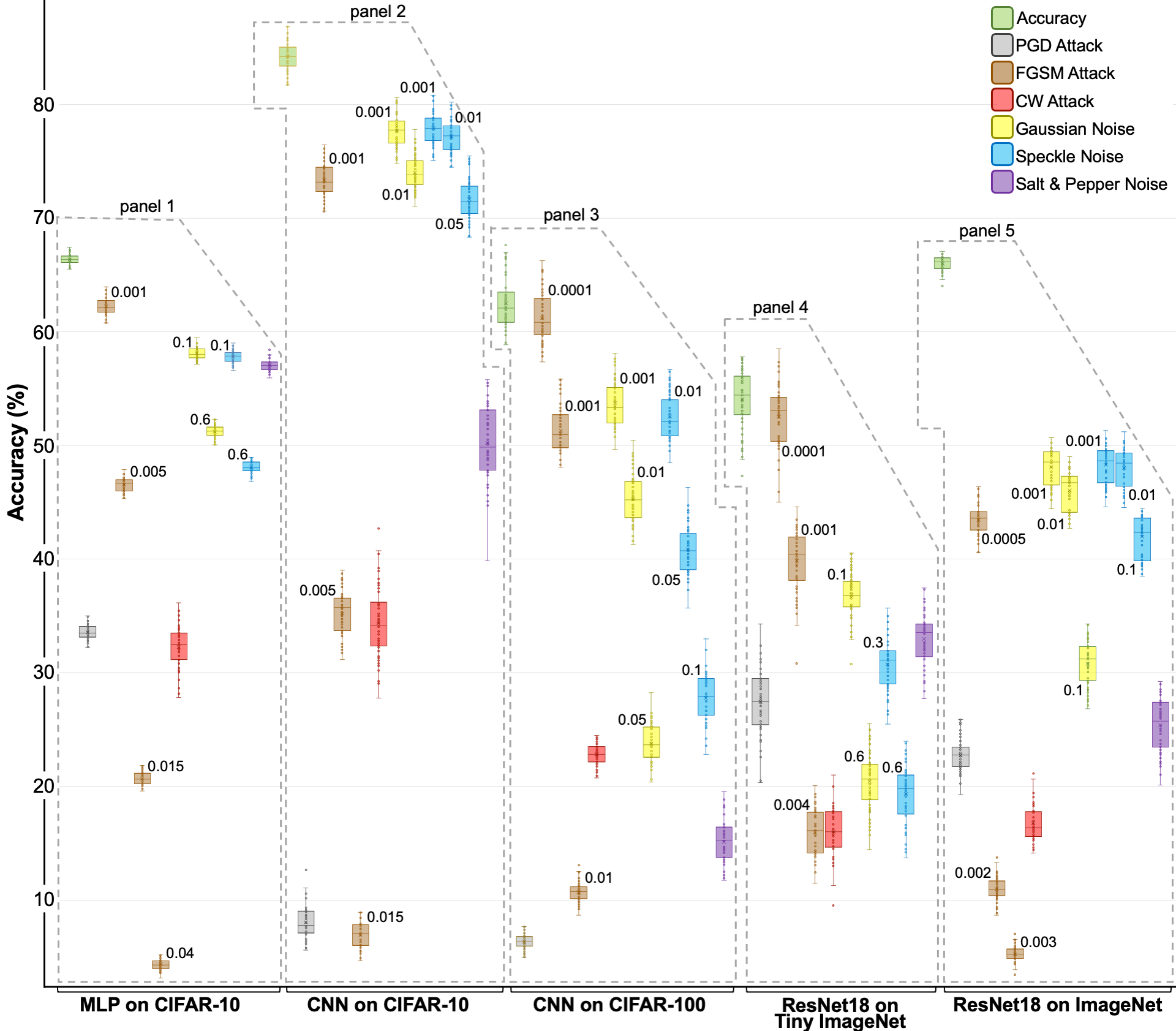}
        \caption{Test accuracies of different DANNs (MLPs, CNNs, and ResNets) trained using image classification datasets (CIFAR-10/100, (Tiny)-ImageNet) and evaluated under noisy conditions are presented. Each box represents 54 different DANNs trained on the indicated dataset. DANNs include (1) 5-layer MLPs trained on CIFAR-10, (2) 8-layer CNNs on CIFAR-10, (3) 8-layer CNNs on CIFAR-100, and (4) ResNet-18 on Tiny ImageNet and (5) ResNet-18 on ImageNet. All DANNs are evaluated using clean validation images, adversarial examples (FGSM, PGD, CW attack), and noisy images (Gaussian, Speckle, Salt\&Pepper noise). The severity levels for FGSM attack, Gaussian and Speckle noise are indicted on respective boxes. We observe a consistent decline in the predictive performance of all DANNs as the severity levels of adversarial attacks or natural noise increase.}\label{fig:box-plot}
        \end{figure}
        
    \subsection{Performance trends of DANNs}\label{subsec23}
    Fig. \ref{fig:box-plot} presents predictive performance of different MLPs, CNNs, and ResNets built using 54 selected graphs and trained on four different image classification datasets. Performance evaluation of the trained DANNs is done using randomly selected 30 different sets of clean, adversarial, and noisy images. The test accuracy numbers presented in Fig. \ref{fig:box-plot} are average values across all tests.
    
        \subsubsection{MLPs on CIFAR-10}
        Panel 1 of Fig. \ref{fig:box-plot} presents test accuracies of 54 MLPs under different conditions. The average clean test accuracy is $66.3\pm0.46\%$, which drops to $33.6\pm0.64\%$ under PGD attack and to $32.3\pm1.9\%$ for the CW attack. With FGSM attack levels of $\epsilon$=[0.001, 0.005, 0.015, 0.04], the test accuracy drops to [$62.2\pm0.72\%$, $46.6\pm0.6\%$, $20.7\pm0.54\%$, $4.34\pm0.44\%$]. For low noise level of natural or additive noise ($\sigma^{2}_{\text{noise}}$=0.1), test accuracy under Gaussian noise is $58.1\pm0.50\%$ and under speckle noise $57.8\pm0.52\%$. For high noise level ($\sigma^{2}_{\text{noise}}$=0.6), the test accuracy  under Gaussian noise is $51.2\pm0.55\%$ and under speckle noise $48.1\pm0.55\%$. Under Salt\&Pepper noise(\emph{salt vs. pepper}=0.5), the test accuracy is $57.06\pm0.5\%$. 
    
        \subsubsection{CNNs on CIFAR-10}
        Panel 2 of Fig. \ref{fig:box-plot} shows the average test accuracies of 8-layer CNNs built from the same 54 candidate graphs. We observe that the average clean test accuracy for CNNs is $84.19\pm1.26\%$, dropping to $8.07\pm1.43\%$ under PGD attack, and to $34.35\pm3.12\%$ under CW attack. We noticed similar trends for various levels of FGSM attacks, as well as for the Gaussian, speckle, salt\&pepper noise. 
        
        \subsubsection{CNNs on CIFAR-100}
        In panel 3 of Fig. \ref{fig:box-plot}, we present CNNs trained on CIFAR-100 dataset. The average test accuracy is $62.49\pm2.24\%$ for clean test dataset, which reduces to $6.35\pm0.64\%$ for the PGD attack, and $22.83\pm0.94\%$ for the CW attack. With FGSM attack levels of $\epsilon$=[0.0001, 0.001, 0.01], the test accuracy is [$61.19\pm2.30\%$, $51.32\pm2.06\%$, $10.71\pm0.83\%$]. For Gaussian noise levels of $\sigma^{2}_{\text{noise}}$=[0.001, 0.01, 0.05], the test accuracy of CNNs is [$53.75\pm2.01\%$, $45.29\pm2.12\%$, $23.76\pm1.66\%$]. For speckle noise levels of $\sigma^{2}_{\text{noise}}$=[0.01, 0.05, 0.1], the test accuracy is [$52.54\pm2.00\%$, $40.82\pm2.11\%$, $27.79\pm2.10\%$]. For salt\&pepper noise, the test accuracy is $15.14\pm1.81\%$. The drop in test accuracy for all cases is significantly more than that of CIFAR-10 dataset.
        
        \subsubsection{ResNet-18 on Tiny ImageNet}
        The panel 4 of Fig. \ref{fig:box-plot} shows 54 different ResNets trained on Tiny ImageNet. The average clean test accuracy is $54.08\pm2.54\%$, $27.45\pm2.90\%$ under PGD attack, and $16.00\pm2.10\%$ under CW attack. For the FGSM attack levels of $\epsilon$=[0.0001, 0.001, 0.004], the accuracy is [$52.53\pm2.85\%$, $39.86\pm2.78\%$, $15.97\pm2.04\%$]. For Gaussian noise levels of $\sigma^{2}_{\text{noise}}$=[0.1, 0.6], the test accuracy is [$36.81\pm1.96\%$, $20.44\pm2.47\%$]. For speckle noise levels of $\sigma^{2}_{\text{noise}}$=[0.3, 0.6], the test accuracy is [$30.73\pm2.31\%$, $19.34\pm2.53\%$]. For salt\&pepper noise, the test accuracy is $33.00\pm2.24\%$.
        
        \subsubsection{ResNet-18 on ImageNet}
        Panel 5 of Fig. \ref{fig:box-plot} presents ResNets trained using ImageNet. Due to the large number of images available for training, the average clean test accuracy of all 54 ResNets-18 was $66.0\pm0.62\%$, a significant improvement over Tiny ImageNet experiments ($54.08\pm2.54\%$). Under PGD attack, the test accuracy drops to $22.75\pm1.39\%$, and to $16.78\pm1.54\%$ under CW attack. For the FGSM attack levels of $\epsilon$=[0.0005, 0.002, 0.003], the accuracies are [$43.47\pm1.40\%$, $10.96\pm1.09\%$, $5.27\pm0.65\%$]. Similar trends are observed for the additive Gaussian and speckle noise under the $\sigma^{2}_{\text{noise}}$=[0.001, 0.01, 0.1]. For salt\&pepper noise, the test accuracy drops to $25.39\pm2.35\%$.
        
        \subsubsection{Comparison of MLPs vs. CNNs on CIFAR-10}
        We observed that CNNs achieve higher accuracy on the clean test data as compared to MLPs on CIFAR-10 dataset. However, under adversarial conditions (FGSM, PGD, and CW attacks), the drop in the performance of CNNs is significantly higher than MLPs as shown in panels 1 and 2 of Fig. \ref{fig:box-plot}. The test accuracy drop is $\sim76\%$ for CNNs compared to $\sim33\%$ for MLPs under PGD attack. For the CW attack, the accuracy drop for CNNs is $\sim50\%$ compared to $\sim34\%$ for MLPs. The same trend was observed for all severity levels of the FGSM attack. Generally, as expected CNNs outperform MLPs under clean test conditions; however, MLPs are more robust to adversarial perturbations as compared to CNNs. We argue that the observed fragility of CNNs is linked to their weight sharing and shift-invariant characteristics, which was previously noted by Zhang et al.~\space\cite{DBLP:conf/icml/Zhang19}.
        
    \subsection{Robustness analysis}\label{subsec24}
\begin{figure}[ht]%
        \centering
        \includegraphics[width=1\textwidth]{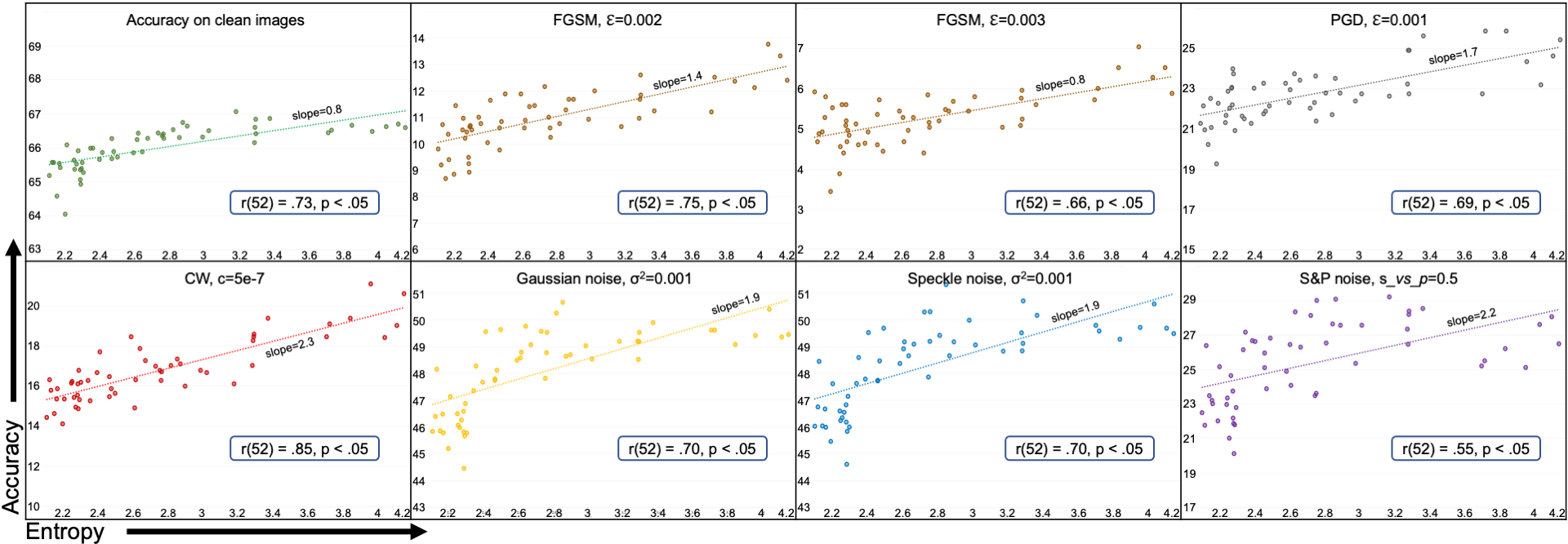}
        \caption{Test accuracy vs. entropy for ResNet-18 on ImageNet. Test accuracy is shown on the vertical axis and entropy (H) of the underlying graph is shown on the horizontal axis. The circles represent an average value calculated over five runs. The type and severity level of noise is shown on the top of each sub-plot. Sub-plots also show trendlines and Pearson correlation coefficients ($r$) with $p$ value. We note significant positive correlation between graph entropy and the performance of the DANNs for all cases.}\label{fig:res18-imagenet}
        \end{figure}

    Our work is a cross-pollination between graph theory and deep learning. We attempt to link the robustness of graphs underlying the architectures of DANNs to their performance against noise and adversarial attacks. On the graph theory side, we use entropy and Ollivier-Ricci curvature to quantify the robustness of graphs. These graphs, in turn, are used to build architectures of DANNs. On the deep learning side, we train these DANNs and quantify their robustness using test accuracy against various types of noise and adversarial attacks. Entropy and Ollivier-Ricci curvature have been extensively studied in the NetSci. These measures have been shown to capture the robustness of cancer networks \cite{tannenbaum2015ricci, sandhu2015graph}, track changes in brain networks caused by age and Autism Spectrum Disorder \cite{farooq2019network}, explain cognitive impairment in patients with Multiple Sclerosis \cite{farooq2020robustness}, identify financial market's fragility \cite{sandhu2016ricci}, and detect communities in complex social networks \cite{sia2019ollivier}. We study the robustness of DANNs and establish the statistical correlation of the observed robustness with entropy and curvature. The correlation results for entropy of graphs and test robustness of DANNs for different datasets are given in Fig. \ref{fig:res18-imagenet}, \ref{fig:res18-tinyimagenet}, and \ref{fig:cnn-cifar100_10}. The correlation results between the robustness of DANNs and graph curvature are provided in Supplementary appendix \ref{secC1}.
    
    \noindent
    \subsubsection{ResNet-18 on ImageNet and Tiny ImageNet}
    Fig. \ref{fig:res18-imagenet} presents 54 ResNet-18 DANNs trained on ImageNet and tested on clean images, adversarial examples generated with FGSM, PGD, and CW attacks, and images with additive Gaussian, speckle, and salt\&pepper noise. Each sub-plot shows entropy (H) of the underlying graph structure and the test accuracy of corresponding ResNet-18 under various conditions. The Pearson product-moment correlation coefficient values between entropy and accuracy along with $p$ values are shown on each sub-plot. There was a positive correlation between the two variables, $r=0.73$, $n=54$, p$<$0.05 for the clean test dataset. We note similar behavior under PGD and CW attacks, that is, a strong correlation between entropy and accuracy exists, $r=0.69$ for PGD and $r=0.85$ for CW, $p<0.05$ for both. Similar trends exist for various severity levels of FGSM attack, Gaussian, speckle, and salt\&pepper noise. In general, across all types of adversarial attacks and noises, the DANNs corresponding to graphs with higher entropy showed stronger robustness and vice versa. Additional results are provided in Supplementary Figs. \ref{fig:supl_ResNet18-ImageNet} and \ref{fig:supl_ResNet18-Tiny-ImageNet}.
    
    \begin{figure}[ht]%
        \centering
        \includegraphics[width=1\textwidth]{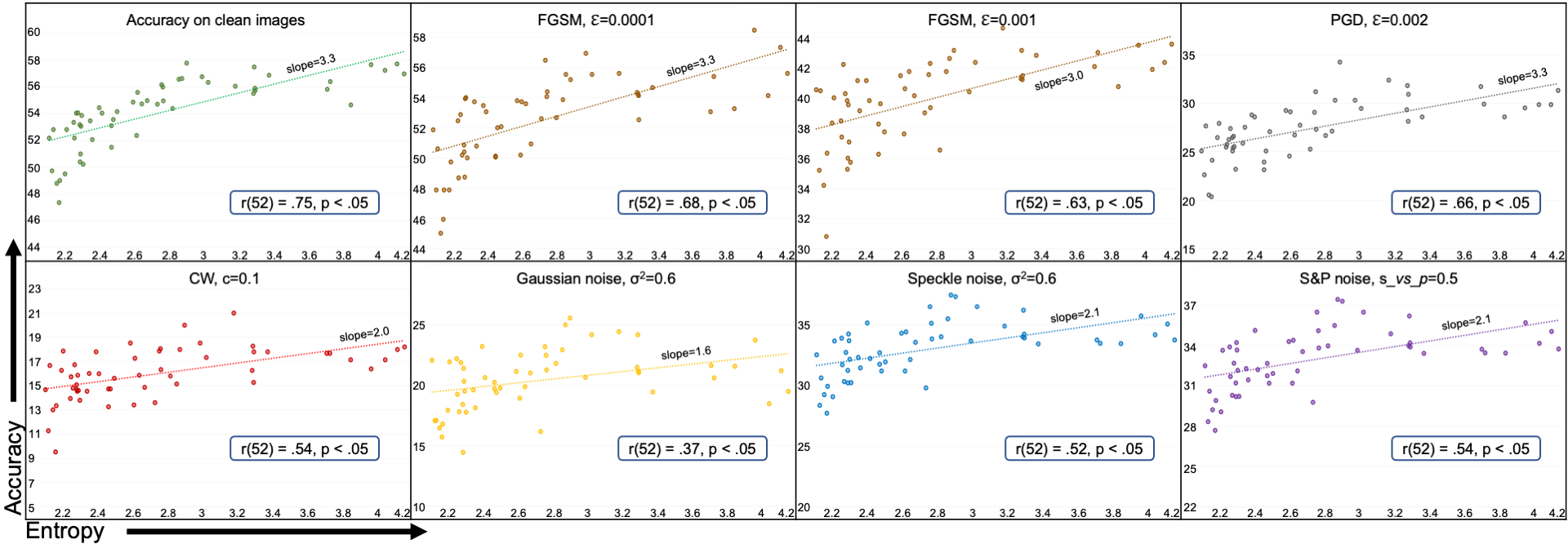}
        \caption{Test accuracy vs. entropy for ResNet-18 on Tiny ImageNet. Test accuracy is shown on the vertical axis and entropy (H) of the underlying graph is shown on the horizontal axis. The circles represent an average value calculated over five runs. We note a strong positive correlation between entropy and the accuracy of DANNs for all types of noise cases.}\label{fig:res18-tinyimagenet}
        \end{figure}
    
    Fig. \ref{fig:res18-tinyimagenet} presents test accuracy vs. entropy plots for 54 ResNet-18 models trained using Tiny ImageNet and tested under various noisy conditions. We observe a strong positive correlation between entropy and predictive performance under all  noise conditions. However, there is a notable decrease in the Pearson product-moment correlation coefficient values in all noise categories compared to the same DANNs when trained and tested on ImageNet. As Tiny ImageNet is a subset of ImageNet with only 200 distinct classes instead of 1,000, the observed decrease in the correlation may be linked to the reduction in complexity of the task, i.e., 200 classes instead of 1,000. 

    \begin{figure}[ht]%
        \centering
        \includegraphics[width=1\textwidth]{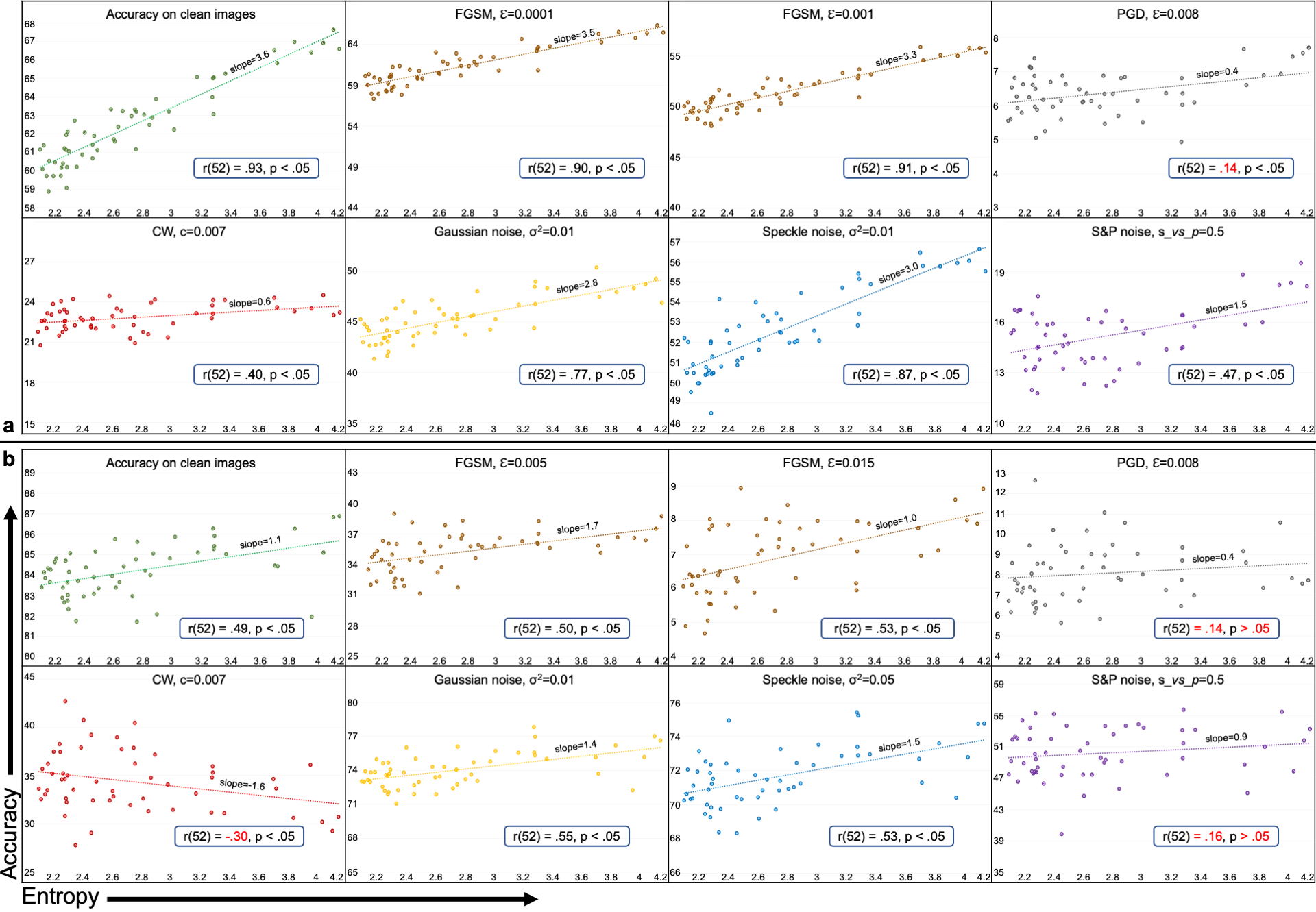}
        \caption{Accuracy vs. entropy for CNNs trained on \textbf{(a)} CIFAR-100 and \textbf{(b)} CIFAR-10 datasets and tested under various noisy conditions. The circles represent average test accuracy over 30 runs. The Pearson correlation $r$ and corresponding $p$ values between entropy and accuracy are also presented for each noise condition. Red text color shows correlation values that are not significant. For the same 8-layer CNNs, the entropy-robustness correlation values increase with the task complexity, that is, relatively higher correlation values are observed for CIFAR-100 as compared to CIFAR-10 dataset for all noise conditions.}\label{fig:cnn-cifar100_10}
        \end{figure}

    \noindent
    \subsubsection{CNNs on CIFAR-100 and CIFAR-10}
    In Fig. \ref{fig:cnn-cifar100_10}(a)\&(b), we present accuracy vs. entropy plots for the 54 8-layer CNNs trained on CIFAR-100 and CIFAR-10 datasets and tested under various noisy conditions. For the CIFAR-100 experiments, we observe relatively strong correlation between entropy and predictive performance except for CW ($r=0.40$, $p<.05$) and PGD ($r=0.14$, $p<.05$) adversarial attacks. For CIFAR-10 dataset, there is a significant correlation between entropy and predictive performance except for the PGD, CW attacks and salt\&pepper noise which were not statistically significant.
    
    We opine that the weak correlation between graph entropy and DANNs' performance under PGD and CW attacks is due to the strong nature of PGD and CW attacks on relatively simple classification tasks of CIFAR compared to Tiny ImageNet and ImageNet. This opinion was strengthened from the evaluation results of the CNNs on a more straightforward classification task of CIFAR-10. We observe that the correlation of entropy with the predictive performance of CNNs reduces for all categories. Moreover, the entropy's correlation with accuracy under CW attack becomes negative. Under PGD attack and salt\&pepper noise, it becomes insignificant with p$>$0.05 as highlighted by the red text in respective subplots of Fig. \ref{fig:cnn-cifar100_10}.
    
    \begin{figure}[ht]%
        \centering
        \includegraphics[width=0.98\textwidth]{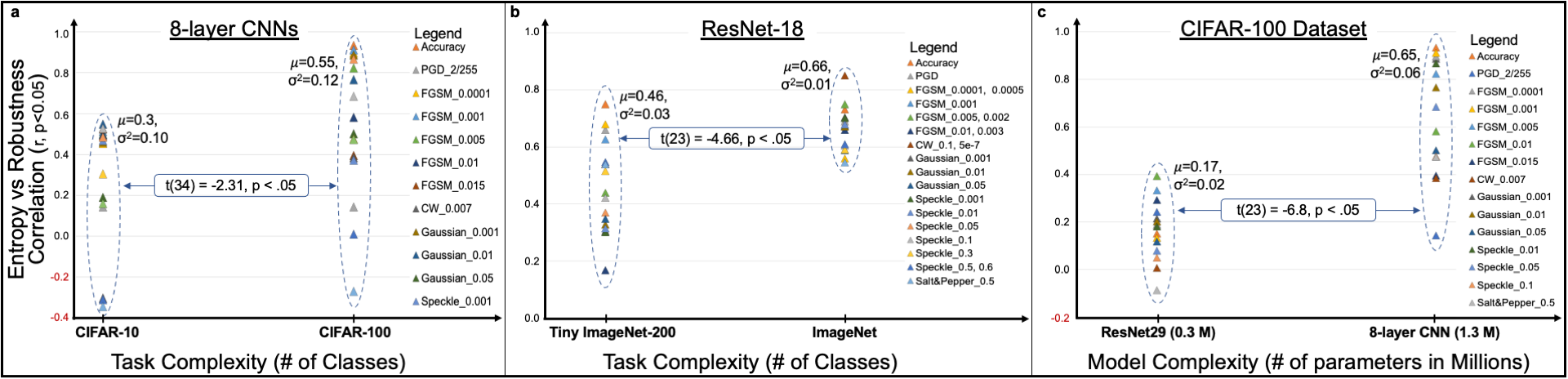}
        \caption{Effect of task and model complexity on \emph{entropy} (calculated in the graph domain) and \emph{robustness} (evaluated in the deep learning domain) relationship. \textbf{(a)}~The entropy-robustness correlation coefficient (vertical axis) is plotted against the number of classes (horizontal axis) for CIFAR-10 and CIFAR-100 datasets for 8-layer CNNs. As the task becomes complex, entropy becomes significantly more correlated to the robustness of DANNs. The inset box shows the Student's t-test statistical analysis. There is significant increase in the entropy-robustness correlation values as the task complexity increases from 10 classes to 100 classes. \textbf{(b)} Entropy-Robustness correlation plot for Tiny ImageNet (200 classes) and ImageNet (1000 classes) datasets. The entropy-robustness correlation for the same DANNs (ResNet-18 in this case) increases significantly as the task becomes complex. ~\textbf{(c)} The entropy-robustness correlation coefficient is plotted against the number of model parameters for different DANNs (ResNet-29 and CNNs) trained and tested using CIFAR-100 dataset. As the number of parameters increases for the same classification task, there is significant increase in the entropy-robustness correlation ($p<.05$).}\label{fig:complexity}
        \end{figure}

    \subsection{Effect of task and model complexity}
    
    We observed that DANNs' robustness, evaluated under noisy conditions, and the robustness of underlying graph structures, quantified using entropy, are strongly correlated. Moreover, this correlation has a strong dependence on the complexity of the model and/or the dataset. In our settings, the model complexity refers to the number of parameters in the model and the task complexity refers to the number of classes in the dataset. As the complexity of the task and/or model increases, the correlation between robust performance and entropy of DANNs increases, as shown in Fig. \ref{fig:complexity}.
    
    In Fig. \ref{fig:complexity}(a), we note that for the same 8-layer CNNs, increasing the complexity of the task (from 10 classes of CIFAR-10 to 100 classes of CIFAR-100) results in increase in the correlation values as noted by the Student's t-test ($t=-2.31,n=34,p<.05$). The same holds true for increasing the task complexity from 200 classes of Tiny ImageNet to 1000 classes of ImageNet and using the same ResNet-18 models ($t=-4.66,n=23,p<.05$), as shown in Fig. \ref{fig:complexity}(b). In Fig. \ref{fig:complexity}(c), we present the effect of increasing the model complexity measured by the number of parameters against the entropy-robustness correlation. We observe that for the same CIFAR-100 dataset, as the model complexity increases from $\sim$0.3 M parameters in ResNet-29 to $\sim$1.3 M in CNN, the entropy-robustness correlation increases significantly ($t=-6.8,n=23,p<.05$).

\section{Discussion}\label{sec12}

In this work, we have shown that graph structural properties such as entropy and curvature can quantify the robustness of DANNs before training. We calculated entropy and curvature of a set of random graphs, which were later transformed into architectures of different types of DANNs. The DANNs were trained and their robustness was evaluated using different types of natural and adversarial noise. We noted that the robustness of trained DANNs was highly correlated with their graph measures of entropy and curvature. We also noted that the said correlations were even stronger for relatively large models and complex tasks.

Currently various autoML and NAS techniques are being developed to search for accurate model architectures for the given datasets and/or tasks. We argue that for many mission-critical applications, the robustness of these models is equally or in some cases more important than accuracy. However, as there are currently no assured ways of estimating the robustness of DANNs in the graph design space except training and testing the candidate DANNs in the deep learning domain. We suggest that the users of autoML/NAS techniques should incorporate entropy and Ollivier-Ricci curvature information into their search framework. Such a practice would allow users or autoML/NAS algorithms choose accurate as well as robust DANNs keeping in view the application area of the machine learning model. The users and autoML/NAS algorithms can identify and choose the most robust model out of all the models that meet the accuracy criteria set by the user. 

A possible future direction is to extend the presented analysis to more complex tasks (e.g., natural language processing) and larger models (e.g., Transformers). Given our current analysis, we anticipate that for the larger datasets, complex tasks, and huge models, the graph robustness measures will be even more relevant and will help users/autoML/NAS algorithms find robust DANN architectures.

\section{Methods}\label{sec11}
We start by presenting the techniques we employed for generating random graphs in the graph theory domain. Next, we describe the graph-theoretical properties used in our experiments to study random graphs. These graph measures are needed to study the structural information of the random graphs. Next, we provide details on transformations for building DANN architectures from random graphs and training these DANNs for various computer vision classification tasks. Finally, we present the multiple conditions, including natural noise and adversarial attacks that we used to evaluate the trained DANNs and quantify their robustness.

    \subsection{Generating Random Graphs}\label{subsec111}
    Random graphs are extensively used in percolation studies, social sciences, brain studies, and deep learning to understand the behavior of natural systems and DANNs \cite{erdHos1960evolution, kang2014random, bassett2006small, bassett2017network, DBLP:conf/iccv/XieKGH19}. We used random graphs, called relational graphs, employed recently in deep learning \cite{DBLP:conf/icml/YouLHX20}.
        
    \subsubsection{Relational graphs}\label{subsec1112}
    A recent a study used relational graphs and showed that the performance of a DANN can be quantified using its graph properties such as clustering coefficient and path length \cite{DBLP:conf/icml/YouLHX20}. 
    The relational graphs are generated through the WS-flex graph generator. WS-flex is a generalized version of the WS model having same-degree constraint relaxed for all nodes. Parameterized by $N$ nodes, $K$ average degree, and $P$ rewiring probability, we represent these graphs by WS-flex$(N,K,P)$. For the graph generator, we use notation $g(\theta,s)$, where $g$ is the generator (for example, WS-flex), $\theta$ represents parameters $(N,K,P)$, and $s$ is the random seed. It is important to note that WS-flex$(N,K,P)$ graph generator encompasses the design space of all the graphs generated by the three classical families of random graph generators, including Watts-Strogatz (WS), Erd{\H{o}}s R{\'e}nyi (ER), and Barab{\'a}si-Albert (BA) \cite{watts1998collective, erdHos1960evolution, albert2002statistical, DBLP:conf/icml/YouLHX20}.

    \subsection{Graph-Theoretic Measures}\label{subsec112}
        
        \noindent 
        \textbf{Average Path Length ($L$)}. It is a global graph measure defined as the average shortest path distance between any pair of graph nodes. It depicts the efficiency of the graph with which information is transferred through the nodes~\cite{mijalkov2017braph}. Small values of $L$ indicate that the graph is globally efficient, and the information is effectively exchanged across the whole network and vice versa. Let $G$ be an unweighted directed graph having $V$, a set of $n$ vertices $\{v_{1}, v_{2}, ..., v_{n}\} \in V$. Let $d(v_{1}, v_{2})$ be the shortest distance between $v_{1}, v_{2}$ and $d(v_{1}, v_{2})$ = 0 if $v_{2}$ is unreachable from $v_{1}$. Then, average path length $L$ is defined as,
            \begin{equation}
                L = \frac{1}{n(n-1)} \sum_{i\neq j}d(v_{i}, v_{j}). \label{eq:L}
            \end{equation}
        
        \noindent
        \textbf{Clustering Coefficient ($C$)}. Clustering coefficient is a measure of the local connectivity of a graph. For a given node $i$ in a graph, the probability that all its neighbors are also neighbors to each other is called clustering coefficient. The more densely interconnected is the neighborhood of a node, the higher is its measure of $C$. Large value of $C$ is linked with the resilience of the network against random network damage~\cite{stam2013connected}. The small-worldness of networks is also assessed by $C$~\cite{DBLP:journals/fini/MasudaSEW18}. For a node $i$ with degree $k_{i}$, clustering coefficient $C_{i}$ is defined as,
            \begin{equation}
                C_{i}=\frac{2d_{i}}{k_{i}(k_{i}-1)},    \hspace{1.5cm}  0 \leq C_{i} \leq 1. \label{eq:C}
            \end{equation}
        where $d_{i}$ is the number of edges between the $k_{i}$ neighbors of node $i$.
        
        \noindent 
        \textbf{Graph Spectral Measures}. The spectral measures focus on eigenvalues and eigenvectors of the associated graph adjacency and Laplacian matrices. We will use topological entropy and Ollivier-Ricci curvature.
        
        \begin{enumerate}
            \item \textbf{Topological Entropy($H$)}. Entropy of graph $G$ having adjacency matrix $A_{G}$, is the logarithm of the spectral radius of $A_{G}$, i.e., logarithm of the maximum of absolute values of the eigenvalues of $A_{G}$~\cite{chen2016robust}.
                \begin{equation}
                    H = \log(\lambda_{A_{G}}). \label{eq:H}
                \end{equation}
            
            \item \textbf{Ollivier–Ricci Curvature (ORC)}. It is the discrete analog of the Ricci curvature~\cite{ollivier2007ricci, ollivier2009ricci}. From the many alternatives of Ricci curvature~\cite{do1992riemannian}, we use the definition presented by Farooq et al.~\cite{farooq2019network} (see Fig. 6 of ref). Let $(X, d)$ be a geodesic (a curve representing the shortest path between two points on a surface or in a Riemannian manifold) metric space having a family of probability measures $\{p{_x}: x \in X\}$. Then, ORC $\kappa_{ORC}(x,y)$ along the geodesic connecting $x$ and $y$ is,
                \begin{equation}
                \kappa_{ORC}(x,y) = 1 - \frac{W_{1}(p_{x}, p_{y})}{d(x,y)},   \label{eq:kappa}
                \end{equation}
            where $W_{1}$ is the earth mover’s distance (Wasserstein-1 metric), and $d$ is the geodesic distance on the space. Curvature is directly proportional to the robustness of the network. The larger the curvature, the faster will be the return to the original state after perturbation. Smaller curvature means slow return, which is also called fragility~\cite{farooq2019network}.
        \end{enumerate}
        
        \noindent 
        \textbf{Robustness ($R$)}. It is the rate at which a dynamic system returns to its original state after perturbation. Fluctuation theorem~\cite{demetrius2013boltzmann} states that, given random perturbations to the network, change in robustness $\Delta{R}$ is positively correlated to change in system entropy $\Delta{H}$,
                \begin{equation}
                \Delta{H} \times \Delta{R} > 0.     \label{eq:H_R}
                \end{equation}
        Entropy $\Delta{H}$ and curvature $\Delta{\kappa_{ORC}}$ are also positively correlated (see Equation (7) of Tannenbaum et al.~\cite{tannenbaum2015ricci}), that is,
                \begin{equation}
                \Delta{H} \times \Delta{\kappa_{ORC}} > 0.      \label{eq:H_kappa}
                \end{equation}
        From Equations (\ref{eq:H_R}) and (\ref{eq:H_kappa}), we see that graph curvature and robustness are also positively correlated,
                \begin{equation}
                \Delta{\kappa_{ORC}} \times \Delta{R} > 0. \label{eq:kappa_R}
                \end{equation}
        Equations (\ref{eq:H_R}) and (\ref{eq:kappa_R}) are the primary motivation in this work to study the curvature and entropy of deep neural networks.

    \subsection{From graphs to DANNs}\label{subsec113}
    Let $G=(V, \varepsilon)$ be a graph having node-set $V=\{v_1, v_2,..., v_n\}$, where node $v$ has feature vector \textbf{x$_{v}$}, and edge set $\varepsilon$=$\{(v_i, v_j)\mid v_i, v_j\in V\}$. The neighborhood of node $v$ is defined as $N(v)=\{u\mid(u,v) \in \varepsilon\}$. To transform the graphs into DANNs, we adopt the concept of  \emph{neural networks as relational graphs}~\cite{DBLP:conf/icml/YouLHX20}. In relational graph, a single \emph{node} represents one input channel and one output channel. \emph{Edge} in the relational graph represents a \emph{message exchange} between the two nodes it connects. The message exchange is a message function having node feature \textbf{x$_{v}$} as input and a message-aggregation function as output. The aggregation function takes a set of messages as input and gives an updated node feature as output. One iteration of this process is one round of message exchange. At each round, each node sends messages to its neighbors, receives messages from all the neighbors, and aggregates them. At each edge, message transformation occurs through a message function $f(.)$, followed by summation at each node through an aggregation function $F(.)$. The $i$-th message exchange round between nodes $v$ and $u$ can be expressed as,
            \begin{equation}
            \mathbf{x}_{v}^{(i+1)} = F^{(i)}(\{f_v^{(i)}(\mathbf{x}_{u}^{(i)}), \forall u \in N(v)\}). \label{eq:msg-exch}
            \end{equation}

    You \emph{et al.} have shown that Equation (\ref{eq:msg-exch}) is the general definition of message exchange that can be used to instantiate any neural architecture~\cite{DBLP:conf/icml/YouLHX20}. We generate MLP, CNN, ResNet-18, and ResNet-29 for each of the 54 random graphs generated from the WS-flex generator.

    The same 54 WS-flex random graphs were transformed into a total of 216 DANNs having 54 neural networks in each of the four categories (MLP, CNN, ResNet-18, and ResNet-29). MLPs were trained on CIFAR-10 dataset, whereas, the CNNs were used for training on CIFAR-10 and CIFAR-100 datasets. The same ResNets-18 were used for training on ImageNet and Tiny ImageNet datasets. The baseline architectures have a complete graph structure for each architecture category. To ensure consistency of our results, we trained each MLP and CNN five times and ResNets one time on respective datasets. The results reported in this paper are average values calculated for thirty different inferences over random test inputs for each MLP and CNN, whereas, five random test inference runs for each ResNet. The compute resources and wall clock times are given in Supplementary appendix \ref{secD1}. List of frameworks and hyperparameters used in our experiments are provided in Supplementary appendix \ref{secE1}.
    
    \subsection{Datasets}\label{subsec114}
    We used four different image classification datasets for our experiments that allowed us to train DANNs of different sizes on tasks that varied in their complexity. We used 10-class CIFAR-10~\cite{krizhevsky2009learning} dataset to train MLPs and CNNs. CIFAR-100~\cite{krizhevsky2009learning} dataset having 100 classes was used to train CNNs and ResNet-29. Both datasets have 50,000 training images and 10,000 validation images. To further scale our experiments, we trained ResNet-18 on the Tiny ImageNet~\cite{TinyImageNet} dataset having 200 classes. Each class in Tiny ImageNet has 500 training images and 50 validation images. We also trained ResNet-18 on the ImageNet~\cite{DBLP:journals/ijcv/RussakovskyDSKS15} dataset having 1,000 classes, 1.2~M training images and 50,000 validation images.
    
    \subsection{Robustness analysis}\label{subsec115}
    We assessed the robustness of DANNs against natural additive noise and malicious noise (adversarial attacks). First, we evaluated the models using clean test images from respective datasets. Then we fed DANNs with different test images corrupted with additive noise and adversarial attacks. It is important to note that we chose the severity levels of adversarial attacks and additive noise so that the predictive performance of DANNs is at the minimum greater than 3\%. We observed at higher levels of noise, the performance would naturally drop to 0\%, which was not helpful in our analysis. Moreover, different severity levels work on different datasets owing to the inherent features and attributes of the data.
        
        \noindent 
        \textbf{Performance evaluation under adversarial attacks}.
        We evaluated DANNs using adversarial examples generated from three different types of attacks, (1) Fast Gradient Sign Method (FGSM)~\cite{goodfellow2014explaining}, (2) Projected Gradient Descent (PGD)~\cite{madry2017towards}, and (3) Carlini Wagner (CW)~\cite{DBLP:conf/sp/Carlini017}. 
        
        Consider a valid input $x_{0}$ and a target class $y_{0}$. It is possible to find $x$ through imperceptible non-random perturbation to $x_0{}$ that changes a DANN's prediction to some other $y$; such $x$ is called an \emph{adversarial example}. Given a loss function $J$($\boldsymbol{x}$;$\boldsymbol{w}$), $\boldsymbol{x}_{0}$ be the input to the model having parameter $\boldsymbol{w}$, the adversarial example $\boldsymbol{x}$ is created by the adversarial attack as,
        \begin{align}
            & \text{FGSM} : &&x = x_{0} + \epsilon \cdot \text{sign}(\nabla_{x}{J(x_{0};w))}, \label{eq:fgsm}\\
            & \text{PGD} : &&x^{t+1} = \Pi_{x+B} \{x^{t} + \alpha \cdot sign(\nabla_{x}{J(x^{t};w)})\}, \label{eq:pgd}\\
            & \text{CW} : &&\underset{x}{\text{min}} \Vert x-x_{0} \Vert^{2} + c \cdot \text{max}\{(\underset{i \neq j}{\text{max}}\{g_{j}(x)\} - g_{t}(x)), 0\}. \label{eq:cw}
        \end{align}

        In Equation (\ref{eq:fgsm}), $\epsilon$ is the severity level of the attack and should be small enough to make the perturbation undetectable. In Equation (\ref{eq:pgd}), $x^{t}$ is an adversarial example after $t$-steps, $\alpha$ is the step-size, $\Pi_{x+B}$ refers to the projection operator for each input $x$ having a set of allowed perturbations $B$ chosen to capture the perceptual similarity between images. In Equation (\ref{eq:cw}), $c>$ 0 is the attack magnitude, $i$ is the input class, and $j$ is the target class. FGSM and PGD have the $l_{\infty}$-distance metric, whereas CW, a regularization-based attack, has $l_{2}$-distance metric in our analysis. 
        
        For the FGSM attacks, we used eighteen severity levels, $\epsilon$=[0.0001, 0.0005, 0.001, 0.0015, 0.002, 0.0025, 0.003, 0.004, 0.005, 0.01, 0.015, 0.02, 0.025, 0.04, 0.045, 0.06, 0.08, 0.3]. For the PGD attacks on CIFAR datasets, we used $\max(B)=0.008$, $\alpha=2/255$, and $t=7$. For the Tiny ImageNet dataset, we used $\max(B)=0.002$, $\alpha=2/255$, and $t=10$, and for the ImageNet dataset, we used $\max(B)=0.001$, $\alpha=2/255$, and $t=10$. For the CW attacks on CIFAR datasets, we used $c=0.007$ and steps$=100$. For the Tiny ImageNet dataset, we used $c=0.01$, steps$=100$, whereas for the ImageNet dataset, we used $c=5e-7$ and steps$=100$.

        \noindent 
        \textbf{Testing under additive noise}.
        We used three different types of noise to generate corrupt images for all the datasets, (1) Gaussian, (2) speckle, and (3) salt\&pepper noise. For each noise type, we used different levels of corruption quantified by the variance and monitored the performance drop. The noise variance used in our experiments for the Gaussian and speckle noise types are $\sigma^{2}=[0.001, 0.01, 0.05, 0.1, 0.2, 0.3, 0.4, 0.5, 0.6]$. For the salt\&pepper noise type, we used the maximum ratio of \emph{salt vs. pepper}=0.5, where salt changes a pixel value to 1 randomly and pepper changes a pixel value to 0 randomly, in the input image. Sample images for each dataset used in our experiments, with noise types and levels are shown in Supplementary appendix \ref{secF1}.

    \subsection{Statistical analysis}\label{subsec116}
        We conducted various statistical tests to ascertain the significance of our analysis. We computed the Pearson product-moment correlation coefficient to assess the relationship between adversarial accuracy and the graph robust structural properties. We also computed the Pearson product-moment correlation coefficient between different structural graph-theoretic measures as shown in Supplementary Fig. \ref{fig:supl_graph_meas}. We used the Student's t-test to establish that average of the correlations between entropy and robustness for two types of datasets as well as two model types are statistically different. This analysis established how entropy is related to the increase in model size and task complexity. The significance level in all these analyses is set to $95\%$.

\newpage

\section*{Supplementary information}
Supplementary information is available as appendices to this paper.

\section*{Acknowledgments}
This work was partly supported by the National Science Foundation Awards ECCS-1903466 and OAC-2008690.

\section*{Data availability}
The datasets used in this study are publicly available on following links: CIFAR-10 and CIFAR-100 (\url{https://www.cs.toronto.edu/~kriz/cifar.html}), Tiny ImageNet (\url{https://www.kaggle.com/c/tiny-imagenet/overview}), and ImageNet (\url{https://www.image-net.org/}).

\section*{Code availability}
For the simulations in deep learning domain, we have used PyTorch machine learning library, primarily developed by Facebook's AI Research lab. The base-code for relational graph experiments (\url{https://github.com/facebookresearch/graph2nn/}) is under the MIT License with copyright(c)Facebook, Inc. and its affiliates. The calculations in graph domain and graph theoretic measures have been coded in Matlab software. Both of these code-packages (in PyTorch and Matlab) are published in the GitHub repository associated with this paper (\url{https://github.com/Waasem/RobDanns}).

\section*{Authors' contributions}
G.R. and H.F. have equal contribution in conceiving the method. A.W. implemented the algorithm and simulations in Python and MATLAB. A.W. wrote the manuscript and Supplementary information. N.C.B and G.R. verified the simulations, methodology, and improved the manuscript. All authors have read and approved the manuscript.

\newpage
\begin{appendices}

\section{Graph to DANN transformation}\label{secA1}%

    \begin{figure}[ht]%
        \centering
        \includegraphics[width=1\textwidth]{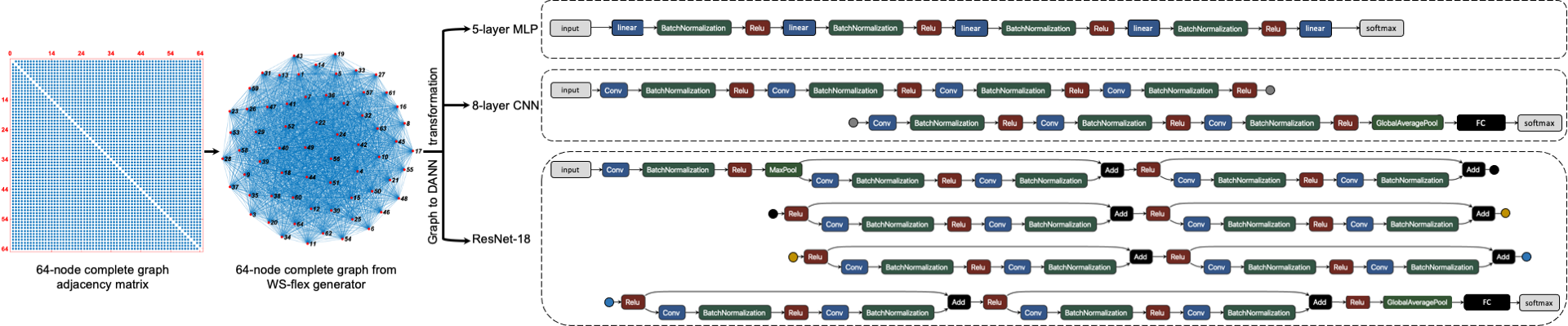}
        \caption{Schematic for graph to DANN transformation. 64-node complete graph generated from the WS-flex generator is shown along with its adjacency matrix. Using \emph{relational graph} transformation, this complete graph is transformed into a 5-layer MLP, 8-layer CNN, and ResNet-18 model. The transformed DANNs are then trained and tested for the given task.}\label{fig:supl_g2nn}
        \end{figure}

\section{Further Results}\label{secB1}%

    \subsection{Limitations of the study}\label{secB11}%
\begin{figure}[ht]%
            \centering
            \includegraphics[width=1\textwidth]{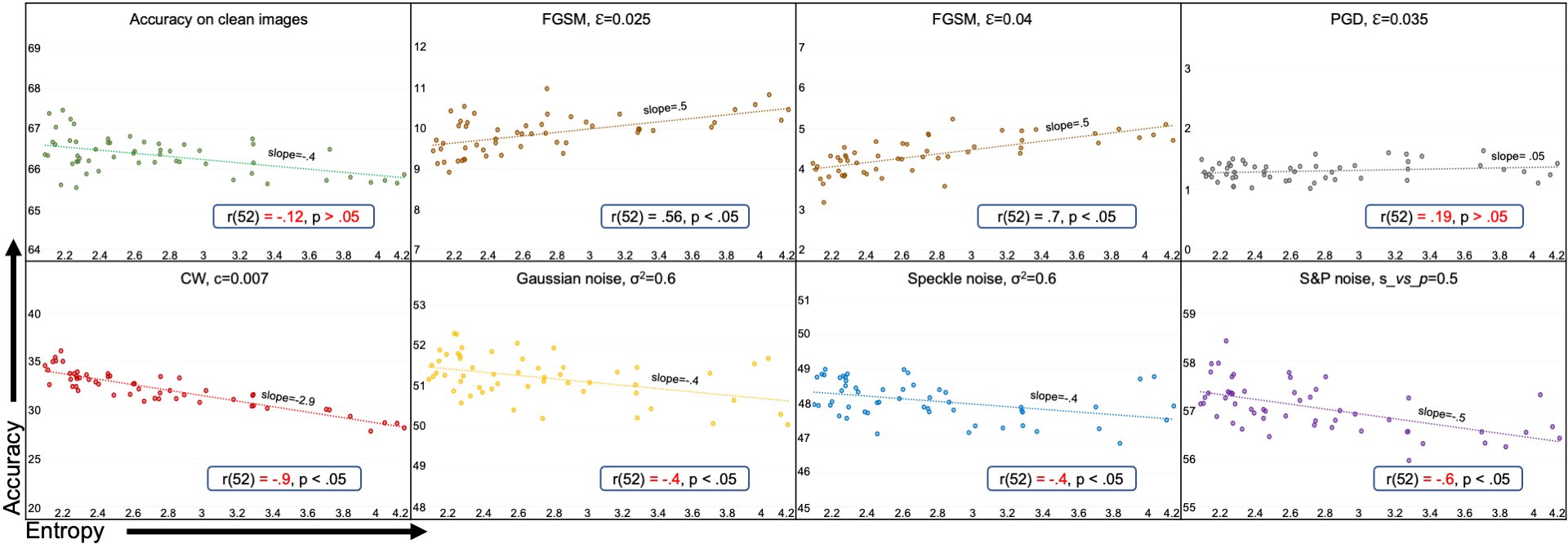}
            \caption{Test accuracy vs. entropy for MLPs trained on CIFAR-10 dataset. The correlations between entropy and test accuracy are insignificant for most of the evaluation categories.}\label{fig:supl_mlp-cifar10}
            \end{figure}

    In our experiments with 5-layer MLPs trained and tested over CIFAR-10 dataset, we noticed that graph structural measures do not efficiently quantify the robustness of MLPs for low severity levels of adversarial attacks and additive noise. MLPs are very dense networks having no weight sharing. Each neuron in MLP has multiple edges across layers, making them fully connected (FC) networks. Under insults such as adversarial attack and natural noise, the MLPs are inherently robust because multiple neurons collectively contribute to the same task. MLPs depict superior accuracy for a given task than CNNs under a robust training regime \cite{driss2017comparison}. We believe that because of the in-built robust nature of MLPs, the graph structural properties such as entropy and curvature do not significantly differentiate robust and fragile MLPs. Results of MLPs trained five times on the CIFAR-10 dataset and randomly evaluated thirty times are illustrated in Fig. \ref{fig:supl_mlp-cifar10}. The correlation between entropy and accuracy of MLPs is insignificant for most of the evaluation categories.
    
        \begin{figure}[ht]%
            \centering
            \includegraphics[width=1\textwidth]{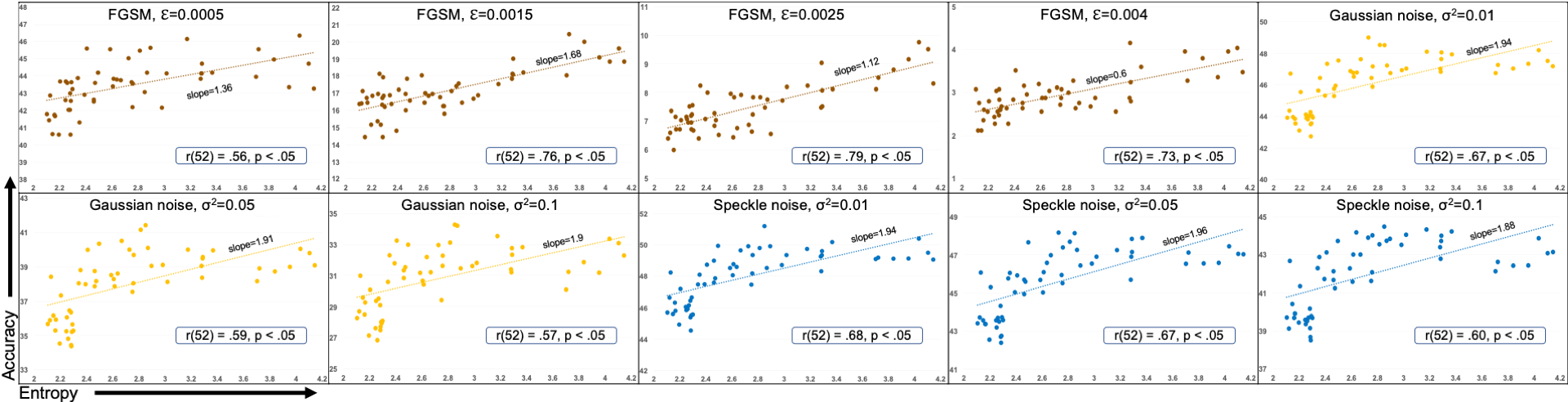}
            \caption{Additional results for ResNet-18 on ImageNet dataset. The experiments show strong positive correlation between entropy and test accuracy for all evaluation categories.}\label{fig:supl_ResNet18-ImageNet}
            \end{figure}
    
        \begin{figure}[ht]%
            \centering
            \includegraphics[width=1\textwidth]{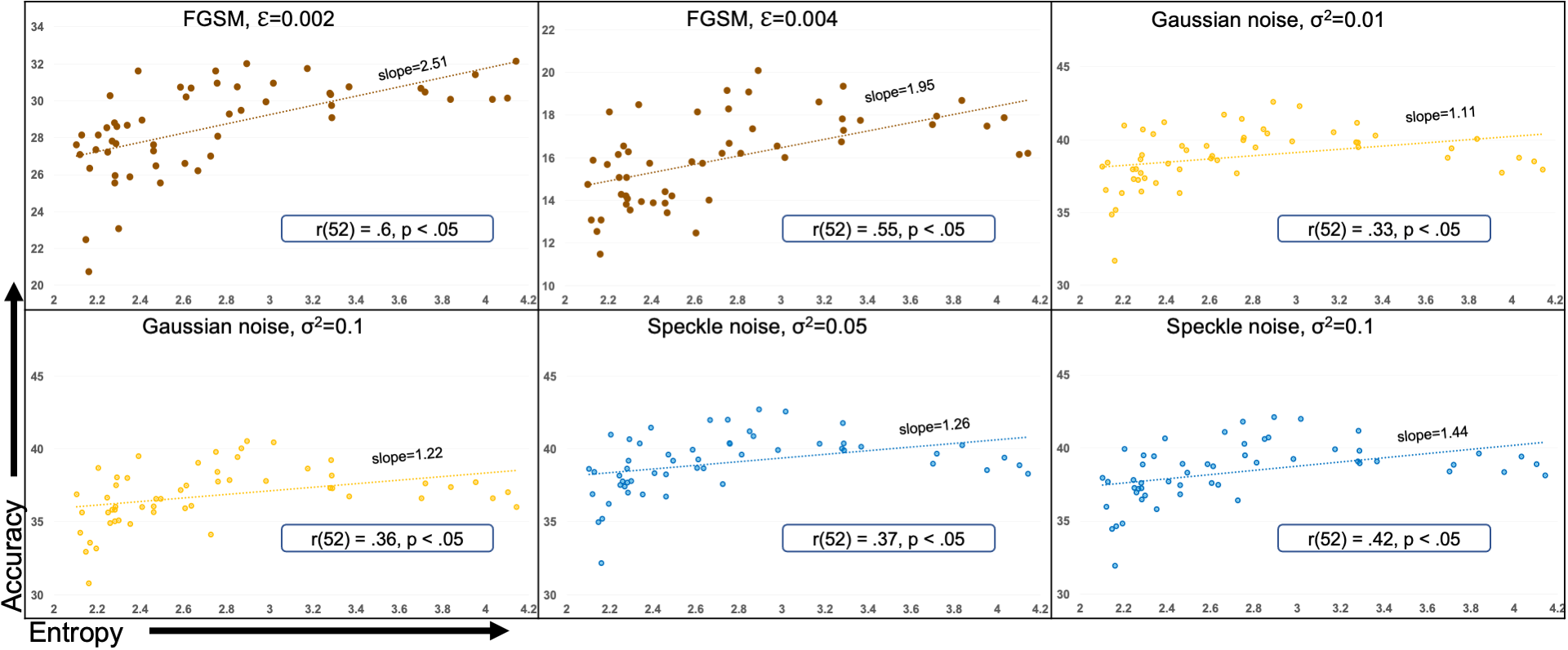}
            \caption{Additional results for ResNet-18 trained and evaluated on Tiny ImageNet dataset.}\label{fig:supl_ResNet18-Tiny-ImageNet}
            \end{figure}
    
        \begin{figure}[ht]%
            \centering
            \includegraphics[width=1\textwidth]{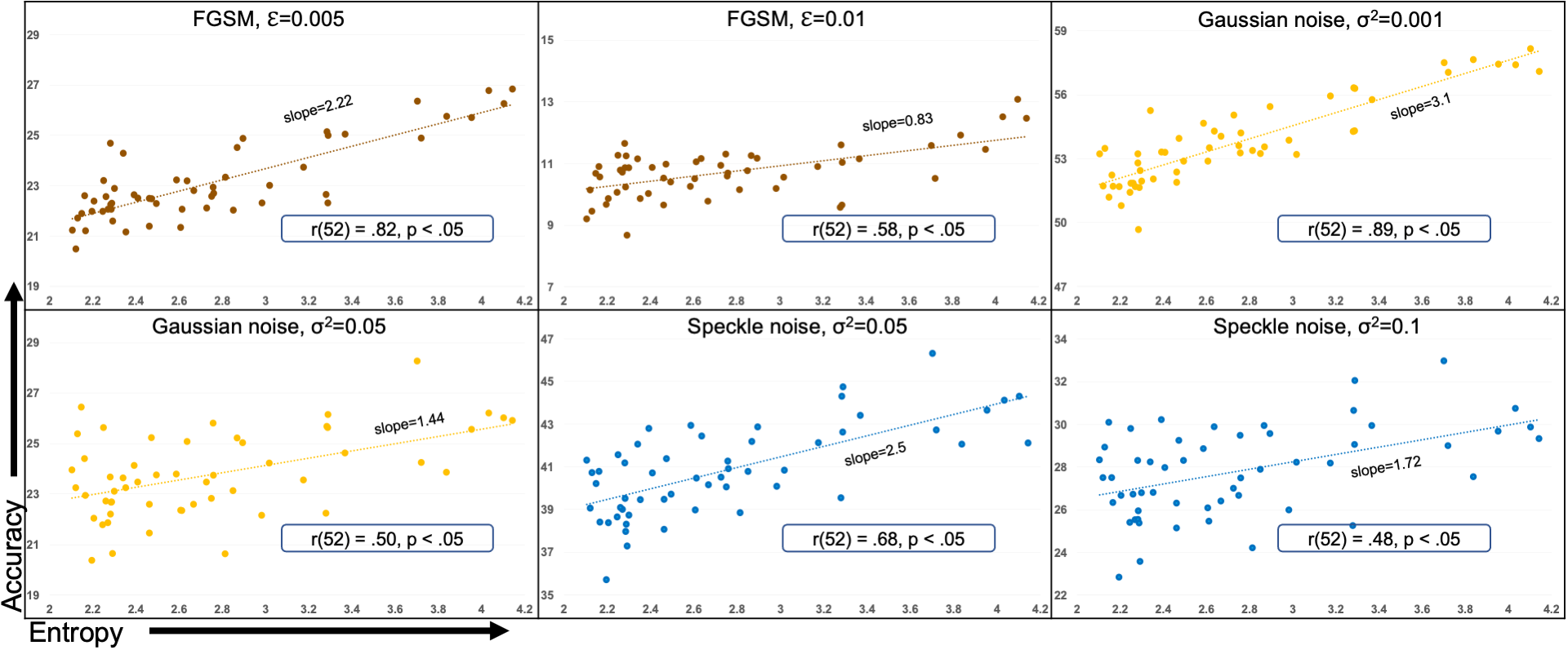}
            \caption{Additional results for robustness evaluation of CNN on CIFAR-100 dataset.}\label{fig:supl_CNN-CIFAR100}
            \end{figure}
    
        \begin{figure}[ht]%
            \centering
            \includegraphics[width=1\textwidth]{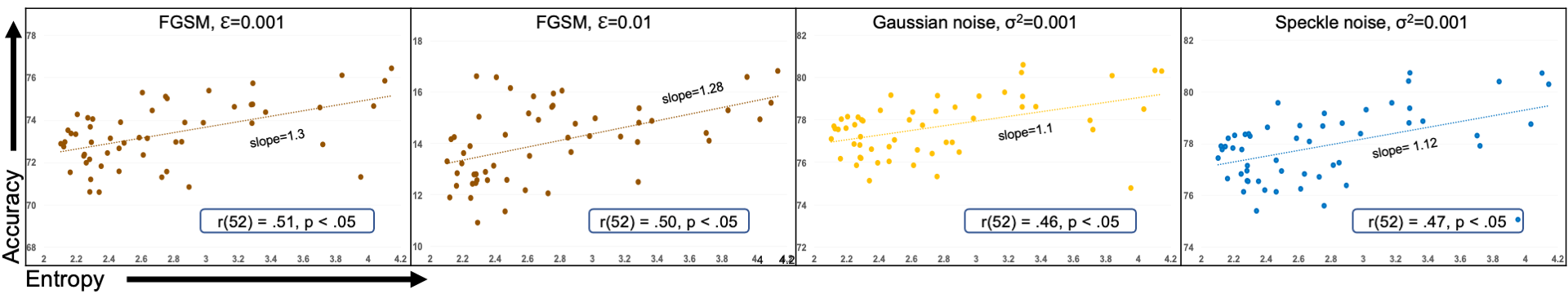}
            \caption{Additional results for robustness evaluation of CNN on CIFAR-10 dataset.}\label{fig:supl_CNN-CIFAR10}
            \end{figure}
        
        \begin{figure}[ht]%
            \centering
            \includegraphics[width=1\textwidth]{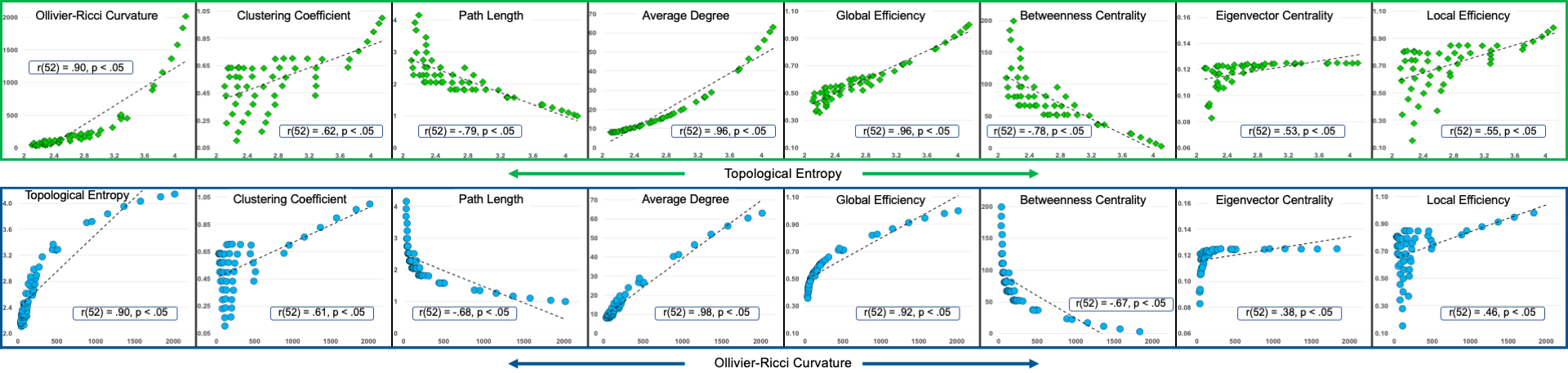}
            \caption{Correlation of Entropy and Curvature with other graph measures considered in our study. The Pearson correlation coefficient is shown in the inset box of each plot. Top row (in green color) shows the correlation of topological entropy with curvature, clustering coefficient, average path length, average degree, global efficiency, betweenness centrality, eigenvector centrality, and local efficiency. The bottom row (in blue color) depicts the correlation between curvature and aforementioned graph properties. We considered various graph-theoretic measures in our experiments for quantifying robustness of DANNs. Consistent with the findings of previous studies in NetSci, the graph structural properties of entropy and curvature are better indicators of DANNs' robustness.}\label{fig:supl_graph_meas}
            \end{figure}
            
\clearpage
\section{Curvature vs. test accuracy}\label{secC1}%

\begin{table}[ht]
\renewcommand {\arraystretch} {1.1}
\begin{center}
\label{tab1}%
\begin{tabular}{l|c c||l|c c}
\hline \hline
      & \multicolumn{2}{c||}{\textcolor{blue}{\textbf{ResNet-18}}} &   &\multicolumn{2}{c}{\textcolor{blue}{\textbf{CNN}}} \\
\hline
\textcolor{blue}{\textbf{Metric}}   &   ImageNet    &    Tiny ImageNet    &     \textcolor{blue}{\textbf{Metric}}  &  CIFAR-100  & CIFAR-10 \\
\hline
Clean Accuracy         &   0.50        & \textbf{0.55} & Clean Accuracy  &   \textbf{0.83} & 0.45 \\
FGSM($\epsilon$=.001) & \textbf{0.56} & 0.49          & FGSM($\epsilon$=.0001) & \textbf{0.79} & 0.42 \\
FGSM($\epsilon$=.002) & \textbf{0.68} & 0.45          & FGSM($\epsilon$=.001) & \textbf{0.83} & 0.46 \\
FGSM($\epsilon$=.003) & \textbf{0.64} & 0.35          & FGSM($\epsilon$=.005) & \textbf{0.81} & 0.43 \\
FGSM($\epsilon$=.004) & \textbf{0.69} & 0.35          & FGSM($\epsilon$=.01) & \textbf{0.64} & 0.44 \\
PGD($B$=0.001) & \textbf{0.61} & -                     & FGSM($\epsilon$=.015) & 0.45 & \textbf{0.46} \\
PGD($B$=.002) & -             & 0.48                  & PGD($B$=.008) & 0.07 & 0.07\textcolor{red}{$^{\dagger}$} \\
CW($c$=5e-7)        & \textbf{0.78} & -                & CW($c$=.007)  & \textbf{0.36} & -0.33 \\
CW($c$=.1)         & -             & 0.38 \\

\hline \hline
        &\multicolumn{2}{c||}{\textcolor{blue}{\textbf{ResNet-18}}} &   &\multicolumn{2}{c}{\textcolor{blue}{\textbf{CNN}}} \\
\hline
\textcolor{blue}{\textbf{Metric}}   &   ImageNet    &    Tiny ImageNet    &     \textcolor{blue}{\textbf{Metric}}  &  CIFAR-100  & CIFAR-10 \\
\hline
Gau($\sigma^{2}$=.001) & \textbf{0.49} & 0.08\textcolor{red}{$^{\dagger}$}  & Gau($\sigma^{2}$=.001) & \textbf{0.80} & 0.41 \\
Gau($\sigma^{2}$=.01) & \textbf{0.45} & 0.10\textcolor{red}{$^{\dagger}$}  & Gau($\sigma^{2}$=.01) & \textbf{0.66} & 0.47 \\
Gau($\sigma^{2}$=.05) & \textbf{0.38} & 0.13\textcolor{red}{$^{\dagger}$}  & Gau($\sigma^{2}$=.05) & \textbf{0.50} & 0.12\textcolor{red}{$^{\dagger}$} \\
Gau($\sigma^{2}$=.1) & \textbf{0.37} & 0.14\textcolor{red}{$^{\dagger}$}  & Spkl($\sigma^{2}$=.01) & \textbf{0.76} & 0.44 \\
Spkl($\sigma^{2}$=.001) & \textbf{0.49} & 0.07\textcolor{red}{$^{\dagger}$}  & Spkl($\sigma^{2}$=.05) & \textbf{0.58} & 0.47 \\
Spkl($\sigma^{2}$=.01) & \textbf{0.46} & 0.09\textcolor{red}{$^{\dagger}$}  & Spkl($\sigma^{2}$=.1) & \textbf{0.43} & 0.23\textcolor{red}{$^{\dagger}$} \\
Spkl($\sigma^{2}$=.05) & \textbf{0.45} & 0.14\textcolor{red}{$^{\dagger}$}  & S\&P($ratio$=.5) & \textbf{0.59} & 0.15\textcolor{red}{$^{\dagger}$} \\
Spkl($\sigma^{2}$=.1) & \textbf{0.36} & 0.21\textcolor{red}{$^{\dagger}$} \\
S\&P($ratio$=.5) & 0.33 & \textbf{0.34} \\

\hline
\end{tabular}
\end{center}
\footnotetext{Pearson correlation coefficient between graph curvature and test accuracy of DANNs. All values except $({\dagger})$ are significant, r(52), p$<$0.05. The bold font indicates the better accuracy of the same DANN on one dataset compared to the other dataset. \textcolor{red}{${\dagger}$} denotes insignificant correlation values, r(52), p$>$0.05. These results indicate that curvature can quantify the robustness of DANNs, especially in complex tasks and bigger models.}
\end{table}

\clearpage
\section{Frameworks and Hyperparameters}\label{secE1}%
Frameworks and corresponding packages used in our experiments are given in Table \ref{tab2}. The hyperparameters used in the training and evaluation of DANNs are given in Table \ref{tab3}. For the sake of procedural consistency and comparisons of results, the set of parameters other than the those mentioned in Table \ref{tab3} are kept the same as in original experiments for relational graphs by their respective authors~\cite{DBLP:conf/icml/YouLHX20}.

\begin{table}[ht]
\renewcommand {\arraystretch} {1}
\begin{center}
\caption{Frameworks and packages used in our codebase.}\label{tab2}%
\begin{tabular}{l||l |c}
\hline 
      & {{\textbf{Package name}}} &   {{\textbf{Version}}} \\
\hline \hline
                                  &   Ubuntu         & 20.04.3 \\
\textbf{Operating systems}        &   Windows        & 10      \\
                                  &   macOS          & 11.6    \\
\hline
\textbf{Programming languages}    &   Python         & 3.6.15  \\
                                  &   Matlab         & R2020a  \\
\hline
\textbf{Deep learning framework}  &   Pytorch         & 1.4.0  \\
                                  &   torchvision     & 0.5.0  \\
\hline
                                  &   RobustBench    & -        \\
\textbf{Adversarial library}      &   torchattacks   & 3.2.1   \\
                                  &   foolbox(optional)   & 3.3.1   \\
                                  &   art(optional)       & 1.9.0   \\
\hline
                                  &   scikit-image   & 0.17.2   \\
                                  &   scikit-learn   & 0.24.2   \\
\textbf{Miscellaneous}            &   scipy          & 1.4.1    \\
                                  &   numpy          & 1.19.5   \\
                                  &   networkx       & 2.3      \\
                                  &   pyyaml         & 5.1.2    \\
\hline
\textbf{Adversarial attacks}      &   FGSM       & - \\
(torchattacks)                    &   PGD        & - \\
                                  &   CW         & - \\
\hline
\textbf{Additive noise}           &   Gaussian        & - \\
(scikit-image)                    &   Speckle         & - \\
                                  &   Salt \& Pepper  & - \\
\hline
\end{tabular}
\end{center}
\end{table}

\begin{sidewaystable}
\sidewaystablefn%
\tiny
\renewcommand {\arraystretch}{1.5}
\begin{center}
\begin{minipage}{\textheight}
\caption{Training and evaluation hyperparameters for our experiments on DANNs.}\label{tab3}%
\begin{tabular*}{\textheight}{@{\extracolsep{\fill}}l||c c|c c|c |c@{\extracolsep{\fill}}}
\toprule%
\hline 
                                & \multicolumn{2}{c|}{{\textbf{CIFAR-10}}} &  \multicolumn{2}{c|}{{\textbf{CIFAR-100}}} & {\textbf{Tiny ImageNet}}  & {\textbf{ImageNet}}   \\
\hline                                
\textbf{Hyperparameter}       & \textbf{MLPs} & \textbf{CNNs}        &  \textbf{CNNs} & \textbf{ResNet-29}    & \textbf{ResNet-18}      & \textbf{ResNet-18}  \\
\hline \hline
Epochs                           & 200             & 100                    &  350             & 150                     & 75                        & 75 \\
\hline
Batch size                       & 256             & 1024                   &  32              & 512                     & 256                       & 450 \\
\hline
Base lr                        & 0.1             & 0.1                   &  0.025             & 0.021                    & 0.1                       & 0.025 \\
\hline
lr policy                       & \multicolumn{5}{c|}{Cosine}                                                                                       & steps=[0, 25, 50, 70] \\
\hline
Momentum                       & \multicolumn{6}{c}{0.9}                                                                                       \\
\hline
Weight decay                  & 0.0005             & 0.01                   &  0.0005             & 0.01                    & 0.006                       & 0.0001 \\
\hline
Drop out                        & -              & -                   &  FC: p=0.1             & -                    & Conv:p=0.2,                     & - \\
                                &                &                     &                        &                      & FC:p=0.5                       & \\
\hline
Trg iterations             & 5              & 5                   &  5                      & 5                    & 1                              & 1 \\
\hline
Eval iterations           & 30             & 30                  &  30                     & 30                   & 5                              & 5 \\
\hline
\end{tabular*}
\end{minipage}
\end{center}
\end{sidewaystable}

\clearpage
\section{Compute Resources and Wall Clock Times}\label{secD1}%
Training time for a 5-layer MLP transformed from the WS-flex random graph on CIFAR-10 dataset is approximately 7 minutes on NVIDIA TITAN RTX GPU. Each MLP was trained five times with random seed, consuming approximately 40 minutes in training the model. On the NVIDIA TITAN RTX GPU, training of all 54 MLPs on CIFAR-10 dataset approximately took 3 days. For CIFAR-100 dataset, the 54 CNNs took approximately 5 days in training the DANNs, five times each. For Tiny ImageNet experiments on the 54 ResNet-18s, the baseline model took approximately 3 hours on TITAN RTX GPU, whereas, the longest training time for a ResNet-18 was approximately 18 hours. Total time for training 54 ResNet-18 models on Tiny ImageNet was approximately 22 days. Training the baseline model of ResNet-18 on ImageNet dataset took approximately 70 hours (3 days) on TITAN RTX GPU, the longest training time for a ResNet-18 model on ImageNet was approximately 123 hours (5 days). Total training time for 54 ResNet-18 models on ImageNet was approximately 3 months with parallel training on four GPUs. All the aforementioned training times include the inference times for FGSM, PGD, and CW adversarial attacks as well as Gaussian, Speckle, and Salt\&Pepper additive noise. For tracking the experiments, visualization of results, and hyperparameter tuning, we used the \emph{Weights and Biases}~\cite{wandb} which is a freely available performance visualization platform for machine learning tasks.

\clearpage
\section{Sample images}\label{secF1}%

    \begin{figure}[ht]%
        \centering
        \includegraphics[width=1\textwidth]{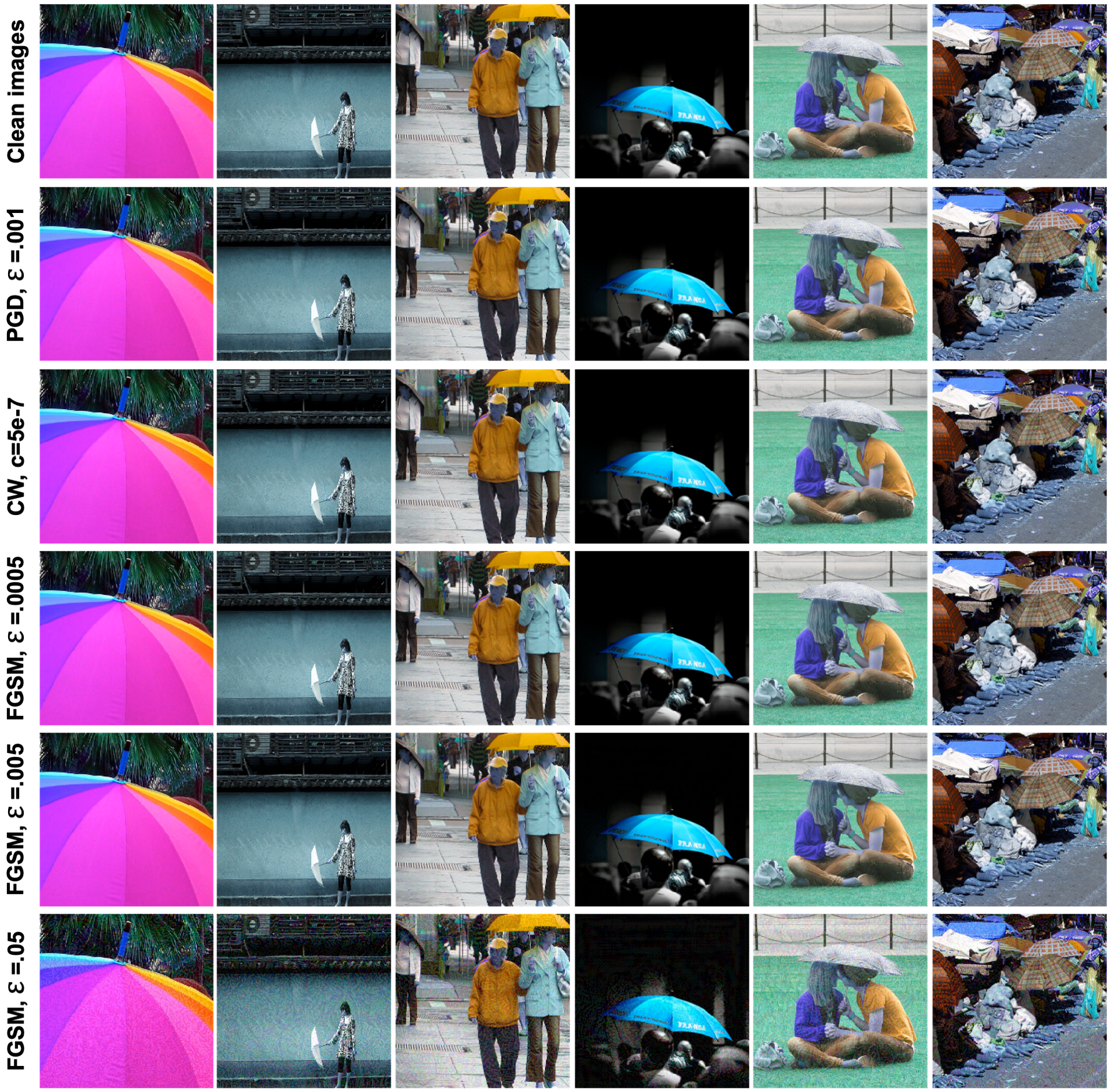}
        \caption{Comparison of clean images from ImageNet dataset with adversarial examples. The severity level for each adversarial attack is shown in the text of the respective row. At higher severity levels such as $FGSM(\epsilon=.05)$, the adversarial noise is noticeable.}\label{fig:supl_adv-imgs}
        \end{figure}

    \begin{figure}[ht]%
        \centering
        \includegraphics[width=1\textwidth]{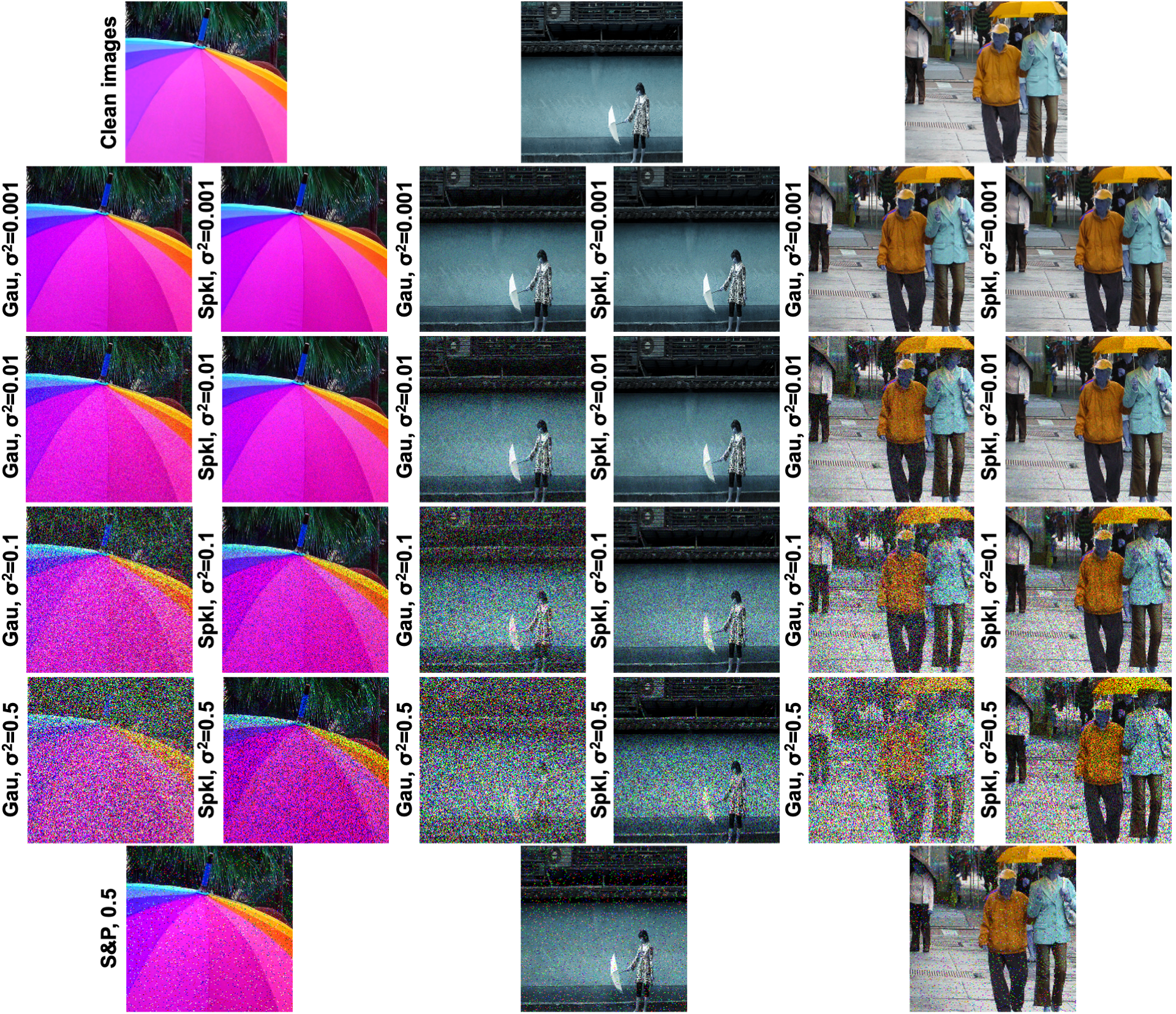}
        \caption{Comparison of clean images from ImageNet dataset with images having natural noise. Three noise types have been used in our experiments; Gaussian, Speckle, and Salt\&Pepper. The severity level for each image is shown as a \emph{variance} ($\sigma^{2}$) for Gaussian and Speckle noise types, and as \emph{salt vs. pepper} ratio=0.5 in Salt\&Pepper noise. As the severity levels increase, the images are visibly distorted.}\label{fig:supl_noisy-imgs}
        \end{figure}
        
\end{appendices}

\clearpage

\bibliography{draft}
\end{document}